\documentclass[11pt]{article} 
\usepackage{soul}
\usepackage[utf8]{inputenc} 
\usepackage{longtable}
\usepackage{url}

\usepackage{geometry} 
\usepackage{graphicx} 

\usepackage{booktabs} 
\usepackage{array} 
\usepackage{paralist} 
\usepackage{verbatim} 
\usepackage{subfig} 
\usepackage[integrals]{wasysym}

\usepackage{fancyhdr} 
\usepackage{sectsty}
\allsectionsfont{\sffamily\mdseries\upshape} 

\usepackage[nottoc,notlof,notlot]{tocbibind} 
\usepackage[titles,subfigure]{tocloft} 

\usepackage{needspace}
\usepackage{natbib}
\usepackage{pdflscape}
\usepackage{afterpage}
\usepackage{chngcntr}
\usepackage{xcolor}
\usepackage{amsfonts}
\usepackage{amsmath}
 
\usepackage[parfill]{parskip} 

\newcommand\ww{.32}
\usepackage[hidelinks]{hyperref} 
\usepackage{longtable}

\newcommand{\basemodl}{llama_instruct_no_chat_l25}

\title{Revealing economic facts: LLMs know more than they say\thanks{Any views expressed are solely those of the authors and so cannot be taken to represent those of the Bank of England or members of its committees. We thank Max Bartolo, Philippe Bracke, David van Dijcke, Caspar Kaiser, Aleksandr Mattal, Paul Robinson, Arzu Uluc and the participants of the CEBRA Annual Meeting 2024 and the ECONDAT 2025 spring meeting.}}
\author{Marcus Buckmann\thanks{Bank of England: marcus.buckmann@bankofengland.co.uk}, Quynh Anh Nguyen\thanks{Bank of England: quynhanh.nguyen@bankofengland.co.uk}, Ed Hill\thanks{Bank of England: ed.hill@bankofengland.co.uk}}

\begin{document}
\maketitle

\begin{abstract}
We investigate whether hidden states of large language models (LLMs) can be used to estimate and impute economic and financial statistics. Focusing on county-level (e.g.\ unemployment) and firm-level (e.g.\ total assets) variables, we show that a linear regression trained on the hidden states of open-source LLMs outperforms the models' own text outputs. This indicates that internal representations encode richer economic information than is revealed directly in generated responses. A learning curve analysis shows that, in many cases, only a few dozen labelled examples suffice for training. We further propose a transfer learning method that improves estimation accuracy without requiring any labelled data for the target variable. Finally, we demonstrate the practical utility of hidden states in data imputation and super-resolution tasks.
\end{abstract}

\clearpage

\section{Introduction} \label{sec:intro}


During training, generative large language models (LLMs) are exposed to vast amounts of information, including data relevant to economic modelling, such as geospatial statistics and firm-level financial metrics. If LLMs can effectively retrieve and utilise this knowledge, they could reduce dependence on external data sources that are time-consuming to access, clean, and merge, or that incur financial costs. Moreover, if LLMs accurately represent data, they could support downstream tasks like data imputation and outlier detection. In this study, we evaluate whether and how LLMs can be used for typical economic data processes.

Not all knowledge within an LLM may be explicit and retrievable in natural language by prompting the model. Instead, we hypothesise that LLMs have latent knowledge about entities such as firms or regions, which is only retrievable with access to the hidden states  -- also referred to as \textit{embeddings} -- of the LLM. For example, suppose we prompt an LLM to estimate the proportion of mortgages in arrears in Orange County, California.  If this exact data point was not present in the training corpus, the model may be unable to provide an accurate written answer, despite having a generalised economic understanding of US counties that could inform an estimate. Furthermore, it is plausible that extensive post-training aimed at reducing hallucinations may have diminished a model’s inclination or ability to make educated guesses. We tap into the LLM's generalised knowledge using its hidden states and find that these allow us to produce better estimates of economic and financial variables than the text output. While proprietary LLMs typically do not expose their hidden states, they are accessible in open-source models.

We test the usefulness of LLMs for data tasks with a large array of empirical analyses on regional economic data of the US, UK, EU, and Germany and financial data on US listed firms. First, we show that a regularised linear model learned on the hidden states often provides substantially better estimates of the economic and financial variables than the text output of the model, particularly for less common statistics. This result holds across open-source LLMs of varying sizes (1--70 billion parameters). A learning curve analysis finds that small samples of labelled data usually suffice to learn an accurate linear model on the embeddings. We also suggest a simple transfer learning algorithm that estimates a statistical variable without \textit{any} labelled data for that variable. We achieve this by both learning from other labelled variables and using the text output of the LLMs as noisy labels on the variable of interest. 

Another way to leverage an LLM’s economic or financial knowledge is by using a reasoning paradigm that allows the model to break down the estimation task into steps, reflect on its knowledge and reasoning process, and refine its approach before arriving at an answer. While we find that a reasoning LLM outperforms the direct text output of a comparable model not tuned for reasoning, our approach—applying a linear model to the prompt’s embeddings—remains superior. It not only delivers better results but is also orders of magnitude more computationally efficient.

Finally, we investigate the practical relevance of our findings for two data processes. Firstly, we show that we can exploit the embeddings to consistently improve the imputation of missing values in economic and financial variables, a common challenge in both research and industrial data pipelines. Secondly, we demonstrate that the hidden states of LLMs can be used for super-resolution tasks, inferring lower-level geographical statistics from high-level data, which can enable the estimation of statistics below the official reporting level. In both applications, we do not assume the LLM has seen the target statistics during training and can recall them by prompting; instead, we leverage the generalised, entity-level knowledge encoded in the model’s hidden states.

Our findings support the value of LLMs in data processing tasks, which warrants attention, given that tasks such as identifying data sources, extracting numeric information, merging data series, and handling missing values are time-consuming and error-prone. This is evident from the numerous data providers that sell packaged, cleaned, and well-structured datasets, even when a large proportion of the underlying data is publicly accessible online. 

This study is structured as follows. Section \ref{sec:lit} discusses the relevant literature, Section \ref{sec:data} describes the datasets and Section \ref{sec:methods} outlines our empirical approach. The main results on the accuracy of our approach to estimate economic variables from embeddings are shown in Section \ref{sec:results_main}. Section \ref{sec:application} presents the methodology and results for two data processing tasks using embeddings: imputation and super-resolution. Section \ref{sec:conclusion} concludes.

\subsection{Literature review} \label{sec:lit}

Our study is most closely related to papers that investigate how well LLMs can recall geospatial statistics. \citet{manvi2023geollm} show that both GPT3.5 and fine-tuned Llama 2 can be prompted to reproduce geospatial information, such as population density, mean income, education, or house prices. The LLMs recall these variables more accurately than supervised machine learning methods, which inferred the information from other data sources such as satellite images of luminosity at night. Note that the authors only consider the text output of the LLM and do not exploit the hidden states as we do. 

Similarly, \citet{li2024can} test how well LLMs can reproduce a large set of variables about cities and regions such as population density, life expectancy, average travel time, and number of patents. As in our study, this paper tests the accuracy of both the text output and a linear model on the hidden layer activations to estimate the statistics. However, they focus on a third approach: they ask the LLM to identify relevant features that can help to predict a variable of interest and then ask the LLM to generate values for these variables. After feeding these variables into a supervised ML model to predict the variable of interest, this approach performs on par with a linear model on the hidden states in 5-fold cross-validation. In contrast to our work, this paper focuses purely on how well the LLMs can recall statistics and does not consider transfer learning or downstream tasks. 

Our main empirical approach of learning a linear model on hidden states of a language model is known as \textit{linear probing} in the literature \citep{belinkov2022probing,alain2016understanding} and has been used to understand the inner workings and capabilities of LLMs.
Our study builds in particular on the findings by \citet{gurnee2023language}, showing that hidden states of LLMs linearly represent space and time. Concretely, a linear model trained on the hidden states of Llama 2 accurately infers the death of historical figures, the release year of artworks, or the geographical location of landmarks. \citet{godey2024scaling} investigate the scaling laws of these results, showing that larger models represent space and time more accurately. Further, \citet{chen2023more} perturb activations in the LLM to show that there is a causal connection between the representations and the text output. 

\citet{zhu2024language} use linear probing to show that LLM embeddings represent the results of simple addition problems. While the exact number cannot be reconstructed using linear probes, the correlation between the actual number output and the inferred number (via the probes) is high. Linear probing is also used to gauge the truthfulness of statements \citep{marks2023geometry, orgad2024llms, burger2024truth} and \citet{liu2024} show that hidden states can also be leveraged to estimate uncertainty of LLM responses in different tasks.

Our work also relates to studies that use hidden states of LLMs for downstream tasks such as classification. Particularly, \citet{buckmann2024logistic} showed how a linear model trained on hidden states of small LLMs after task-specific prompting can outperform GPT-4 when trained on only a few dozen samples per class \citep[see also][]{cho2023palp}. Hidden states of LLMs have also proven to perform well as general text embeddings \citep{lee2024nv,SFRAIResearch2024} as evidenced by their competitive performance on the Massive Text Embedding Benchmark (MTEB) \citep{muennighoff2022mteb}, which includes tasks such as clustering, re-ranking, retrieval and classification. 

Our transfer learning strategy draws on the literature on deep learning from noisy labels \citep[see][for a review]{song2022learning}. Consistent with prior findings that deep networks can generalise beyond noisy supervision and achieve test accuracy above the quality of the training labels \citep{rolnick2017deep,oyen2022robustness}, we observe that our model outperforms the noisy labels derived from LLM outputs. Moreover, following evidence that neural networks learn clean signal early and only later memorise noise, we employ early stopping to curb overfitting \citep{liu2020early}.

The literature on the use of LLMs for data imputation and super-resolution tasks -- the two practical applications of our methodology that we explore in this study -- is scarce. \citet{hayat2024claimdata} fine-tuned Llama 2 to impute missing values given a textual description of the instance's features. \citet{ding2024semantic} use a similar approach for data imputation in recommender systems. By contrast, we do not rely on text output but use the embeddings of the LLM as additional features when employing standard imputation methods. 

While there exists a rich statistical and machine learning literature on the estimation of geospatial statistics at a granular level \citep{tzavidis2018start, chi2022microestimates, aiken2022machine, viljanen2022machine}, we are not aware of any study that explores the use of LLMs for this purpose. 




\section{Methodology} \label{sec:methods}

\subsection{Datasets} \label{sec:data}

We use four public regional datasets covering US counties, EU regions, UK districts and German districts. Additionally, we use a dataset containing public financial information about US listed firms. The datasets are listed in Table \ref{tab:data_description}. We assembled the datasets using different public sources, which are described in detail in Appendix \ref{appx:data}. All datasets are cross-sectional and are based on statistics from 2019, with the exception of the US firms dataset for which we obtain data from 2022.\footnote{The free tier of the Yahoo Finance API limits access to earlier history.} Intentionally, all datasets predate the 2023 knowledge cut-offs of the LLMs we tested so that we can assess whether the LLMs can exactly recall statistics.

\begin{table}[tb]
\begin{small}
\begin{center}
\addtolength{\tabcolsep}{-2pt}

\begin{tabular}{llllll}
\hline
Name & Observations & Variables & Regional level &  Grouping variable & Year \\ \hline
US counties & 3142 & 9 & counties & 51 states & 2019 \\
German districts* & 401 & 13 & districts & 38 gov.\ districts & 2019 \\
UK districts & 374 & 8 & districts (LADs) & 12 regions & 2019\\
EU regions (NUTS2)& 244& 7 & NUTS-2 & 27 countries & 2019 \\
US listed firms & 1986 & 9 & N/A & N/A & 2022 \\

\hline 
\multicolumn{6}{l}{\footnotesize*In German, the regional level is referred to as \textit{Kreise und}} \\
\multicolumn{6}{l}{\footnotesize \textit{ kreisfreie Städte} and the grouping level as \textit{Regierungsbezirke}.}

\end{tabular}

	\caption{Datasets. Appendix \ref{appx:data} provides full documentation of datasets and sources.}
\label{tab:data_description}

\addtolength{\tabcolsep}{2pt}
\end{center}
\end{small}
\end{table}

Key variables that we use in the regional datasets are population, GDP per capita, unemployment rate, a measure of the income per capita, and a measure of life expectancy. Key variables we observe in the US firms dataset are total assets, market capitalisation, profitability metrics such as return on equity and return on assets. The full list of variables in all datasets are shown in Table \ref{tab:tab_cv_all}.  For the super-resolution analysis in Section \ref{sec:super_res} we use additional data that we describe in that Section. We provide more extensive results for our main datasets, US counties and US listed firms, which contain a larger number of observations than the other datasets. 

\subsection{Experimental set-up} \label{sec:setup}

We exclusively use open-source LLMs, as, to the best of our knowledge, none of the providers of the leading proprietary models provide access to the hidden states of their generative models. Furthermore, we focus on smaller models because these are easier and cheaper to use. The models we test are listed in Table \ref{tab:models}. In the main experiments, we consider three models of the Llama 3 family (1B, 8B and 70B parameters) and the Phi-3-mini (3.8B parameters) model. Apart from Llama 3 70B, we accessed the models through the \texttt{transformers} library \citep{wolf2020transformers} on an NVIDIA Tesla V100 GPU. Due to its large size, we use a quantised version of Llama 3 70B on 2 NVIDIA T4 GPUs and obtain the embeddings with the \texttt{llama-cpp-python} library. To obtain this model's text outputs, we use the non-quantised model hosted on \texttt{\href{https://replicate.com/}{replicate.com}}. More details on the computational requirements can be found in Appendix \ref{appx:computation_time}.

We also consider a reasoning model: Qwen QwQ (32B parameters). To assess whether the reasoning paradigm improves the estimation of regional and financial statistics, we compare the model directly against Qwen 2.5 32B, which is the general-purpose LLM on which Qwen QwQ is based. To obtain the text output of both models, we use the instances hosted on \texttt{\href{https://groq.com/}{groq.com}}. We obtain the embeddings of Qwen 2.5 32B from a quantised version using \texttt{llama-cpp-python}.

\addtolength{\tabcolsep}{-2pt}
\begin{table}
\begin{small}
\begin{center}
\begin{tabular}{lll}
Name  & Source & Quantisation \\ \hline 
\textbf{Main models} \\
Llama 3.2 1B-Instruct & huggingface.co/meta-llama & no\\
Phi-3-Mini-4K-Instruct (3.8B)  & huggingface.co/microsoft & no\\
Llama 3 8B-Instruct& huggingface.co/meta-llama & no \\
Llama 3 70B-Instruct & Text: replicate.com/meta & no \\
 & Embeddings: huggingface.co/bartowski & IQ2\_M.gguf \\
 \textbf{Reasoning experiments} & \\
 Qwen QwQ-32B & Text: groq.com: qwen-qwq-32b \\
 Qwen 2.5 32B & Text:  groq.com: qwen-2.5-32b \\
 & Embeddings: huggingface.co/Qwen &  q4\_k\_m.gguf \\

\hline
\end{tabular}
\end{center}
\caption{LLMs used in this study.}
\label{tab:models}
\end{small}
\end{table}
\addtolength{\tabcolsep}{2pt}
 
We prompt the models using a \textit{completion prompt} as follows:

\textit{The \{\texttt{variable}\} in \{\texttt{region}\} in \{\texttt{year}\} was} 

For example, we use the prompt: ``\textit{The population in Orange County, California in 2019 was}". We also tested several other prompting strategies including a question-answering prompt, a few-shot prompt, and a chain-of-thought prompt. Appendix \ref{appx:prompting} shows how the completion prompt delivered the most accurate results, on average.

When training a linear model on the embeddings, we also test a \textit{generic prompt}, which does not vary with the variable of interest but just embeds the name of the entity and the year (e.g.\ \textit{``Orange County, California in 2019"}).

\subsection{Linear model on embeddings (LME)}

We fit a ridge regression model on the hidden states of the prompt's last token. Our baseline model has 32 layers, with a 4096-dimensional embedding vector. If not stated otherwise, we use the embeddings of the 25\textsuperscript{th} layer because LME performs better on this layer than on others (Figure \ref{fig:by_layer} in the appendix).

To estimate the performance of LME, we employ \textit{grouped} cross-validation on the regional datasets. Specifically, we ensure that the training and test sets do not share any values of the grouping variable. The grouping variable for each dataset is shown in Table \ref{tab:data_description}. For US counties, for example, the grouping variable is US states. Our baseline results are based on 25 repeats of repeated 5-fold cross-validation but we also report learning curves, where, adhering to the grouping, we sample training sets of increasing size.\footnote{We repeat the learning curve sampling procedure between 30 and 300 times, depending on the dataset size and training sample size.}

We either train the linear model on the embeddings directly or on the first $k$ PCA components, with $k \in \{5,10,25,50,100, 200 \}$. We fit the PCA to all observations and not only on the training set. 
The regularisation parameter $\alpha$ is tuned over 50 logarithmically spaced candidates between $10^{-5}$ and $10^{5}$. The selected $\alpha$ minimises the mean squared error under 5-fold grouped cross-validation within the training set.

We denote the ground-truth values of the variable of interest by $y$, the embedding matrix by $E$, and the LLM’s textual output by $\overset{txt}{y}$.

\subsection{Data transformations and performance metrics}

To avoid a strong influence of outliers on the performance metrics and on the estimation of LME, we transform some of the dependent variables. Specifically, across all datasets, we applied a log transformation to non-negative variables with a skewed distribution. For skewed variables with negative values, we use a cubic transformation ($f(x) = sgn(x) \times |x|^\frac{1}{3}$). The transformations applied to each variable are reported in Table \ref{tab:tab_cv_all}. 

When assessing the performance of the text output and LME, we test both models on exactly the same observations. The text outputs of the LLMs sometimes fall outside the range of expected values (e.g. unemployment rate~$>50$\%). In this case, we decide to remove the observations. Parsing numeric values from LLMs' textual answers is challenging, due to the inconsistent format and inconsistent use of units. We made considerable effort to correctly parse the text output and conducted comprehensive manual checks on all variables to ensure the accuracy of our parsing approach. Our parsing approach is described in detail in Appendix \ref{appx:number_extract}.

Despite the use of data transformations and the removal of obvious outliers, we still observe some outliers in the values parsed from the text output. To minimise the influence of these outliers on performance, we employ Spearman correlation as our main performance metric, measuring the rank correlation between the ground truth and estimated values of the variable of interest. We also report our key results for Pearson correlation and do not observe a qualitative difference in our conclusions.

\section{Using LLM embeddings to infer statistics} \label{sec:results_main}

We start our analyses with a cross-validation exercise comparing the performance of LME and text output. Next, we run a learning curve analysis to understand how many training samples are required for LME to perform well. Finally, we test different transfer learning approaches, where we try to estimate a variable without access to any labelled data on that variable. In this section, we focus on the performance of the \textit{baseline LME}, a ridge regression model trained on all 4096 embedding dimensions (i.e. without PCA) of the 25\textsuperscript{th} layer of Llama 3 8B.

\renewcommand\ww{.49}

\begin{figure}[!htp]
 \includegraphics[width=\ww\linewidth]{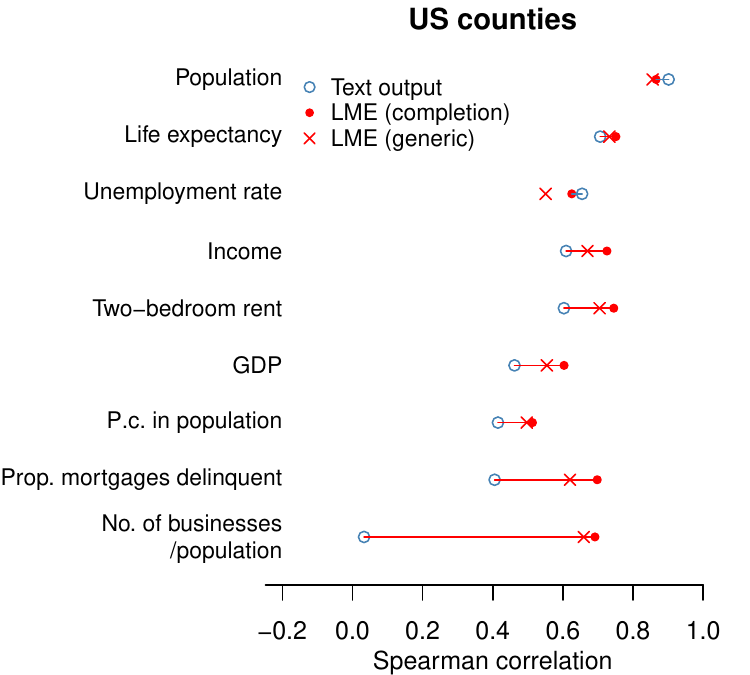} 
 \includegraphics[width=\ww\linewidth]{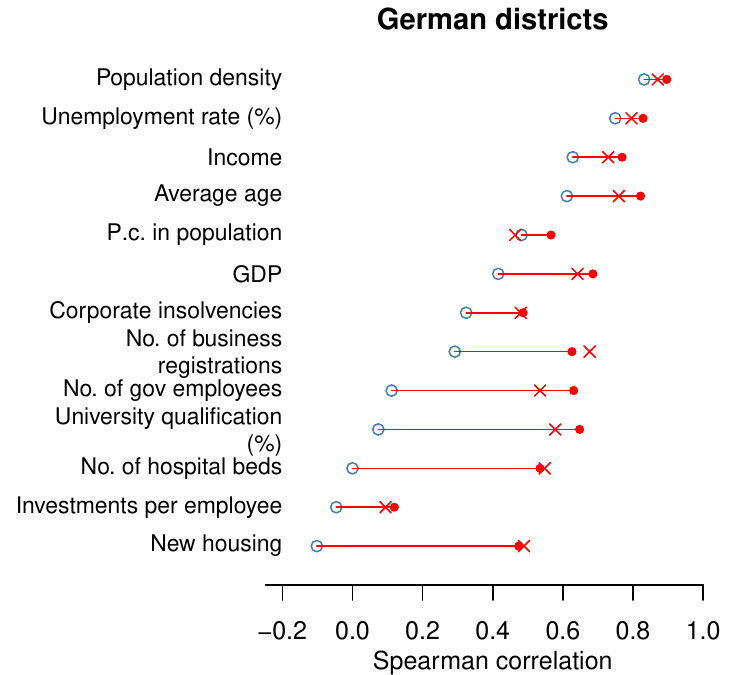} \\
 \includegraphics[width=\ww\linewidth]{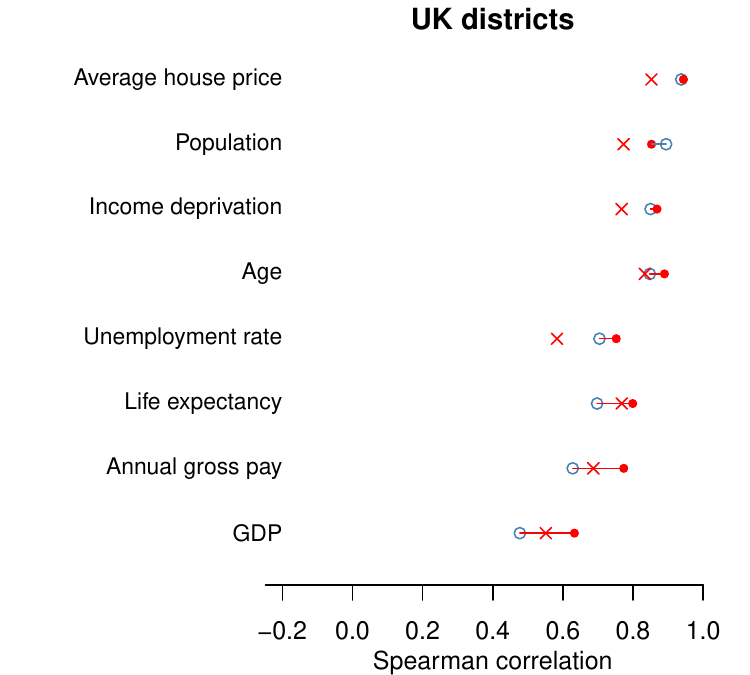}
  \includegraphics[width=\ww\linewidth]{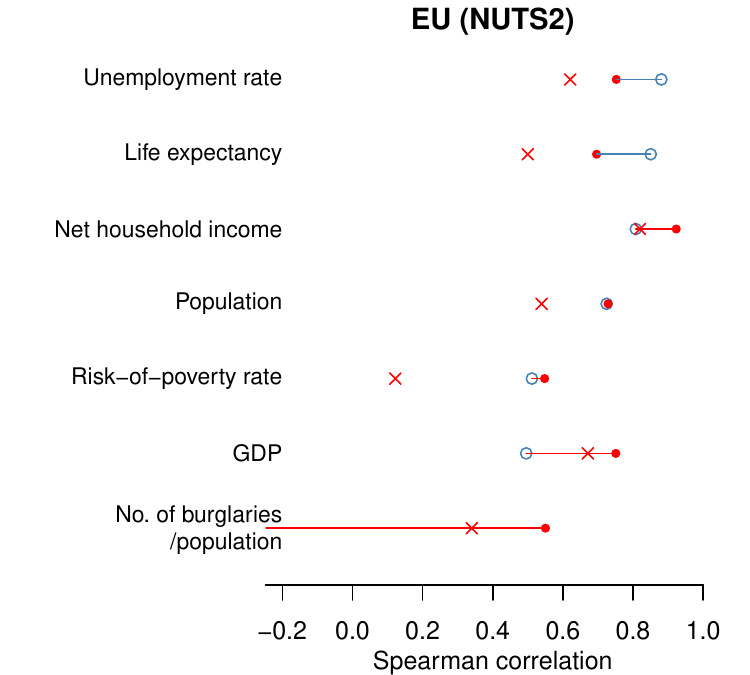}
  \includegraphics[width=\ww\linewidth]{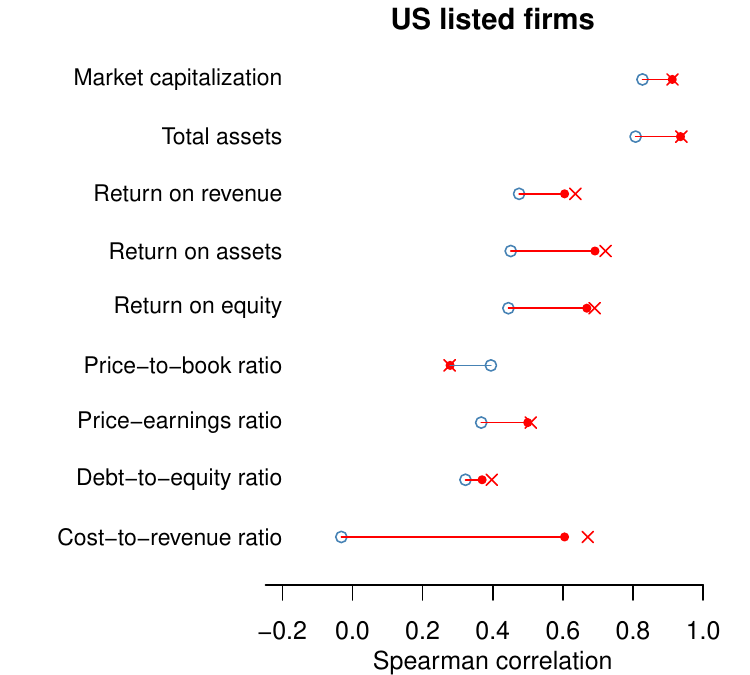}
\caption{Cross-validation performance. The variable labels are shortened. The full variable names used when querying the LLM are shown in Table \ref{tab:tab_cv_all}.}
\label{fig:bars}
\end{figure}

\subsection{Cross-validation}

Figure \ref{fig:bars} compares the performance (Spearman correlation) of the text output to the baseline LME learned on the embeddings of the completion prompt and the generic prompt. The variables are ordered by decreasing performance of the text output.\footnote{See Table \ref{tab:tab_cv_all} for a tabular representation of the results including a measure of uncertainty, which we omitted from these charts to improve legibility.} LME based on the completion prompt performs better than LME on the generic prompt in all datasets except the US firms data, where both perform equally well on most variables. We therefore focus on the completion prompt in the following analyses.

LME beats the text output most of the time. The performance advantage of LME increases with decreasing performance of the text output. On more common variables, such as population, or unemployment rate, the text output performs as well as, or better than, LME. On less common statistics, however, such as the proportion of mortgages that are at least 90 days delinquent (US counties) or the number of government employees (German districts), LME performs substantially better. The results are similar when choosing Pearson correlation as the performance metric, as shown in Table \ref{tab:tab_cv_all} and Figure \ref{fig:bars_pearson} in the Appendix.


\renewcommand\ww{.32}

\begin{figure}[h]

\includegraphics[width=\ww\linewidth]{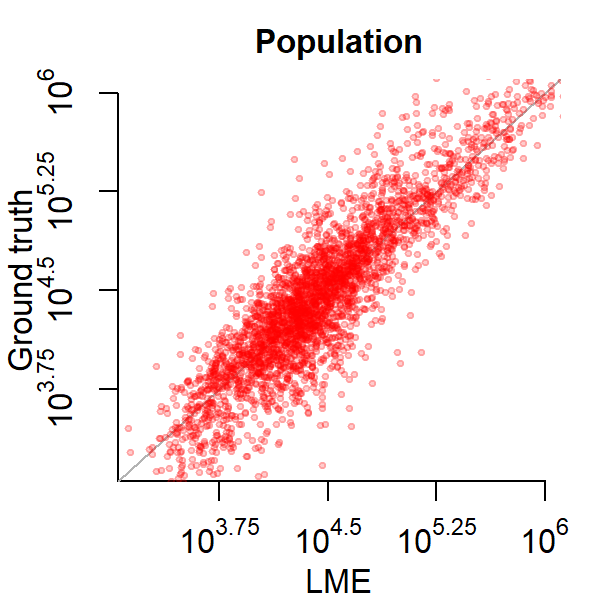}    
\includegraphics[width=\ww\linewidth]{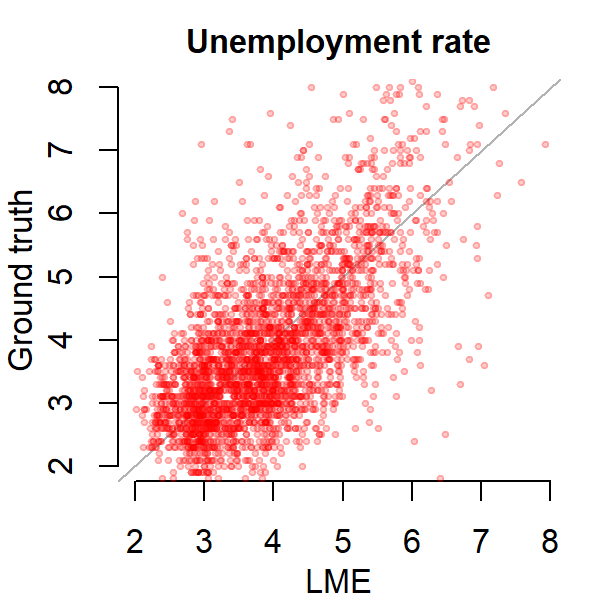}
\includegraphics[width=\ww\linewidth]{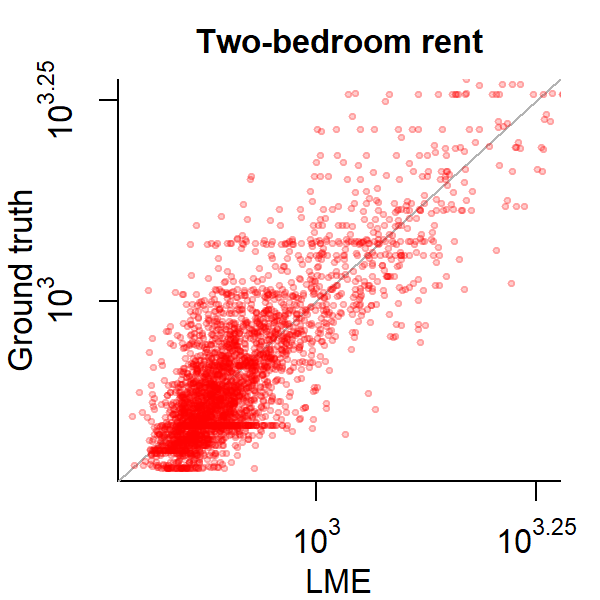}

\includegraphics[width=\ww\linewidth]{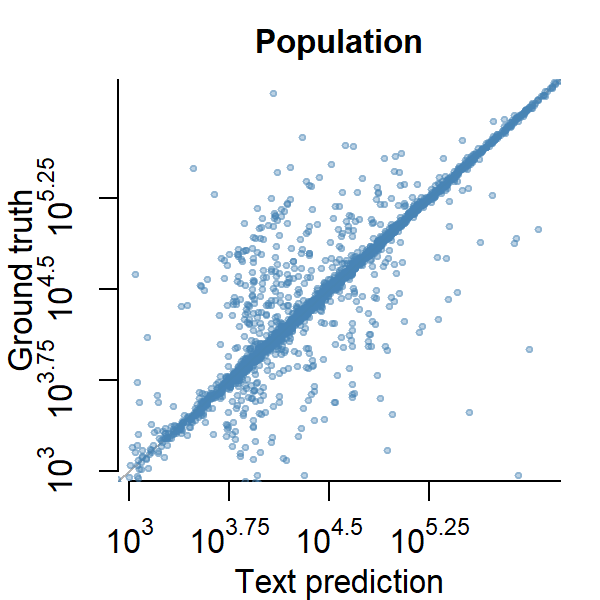}    
\includegraphics[width=\ww\linewidth]{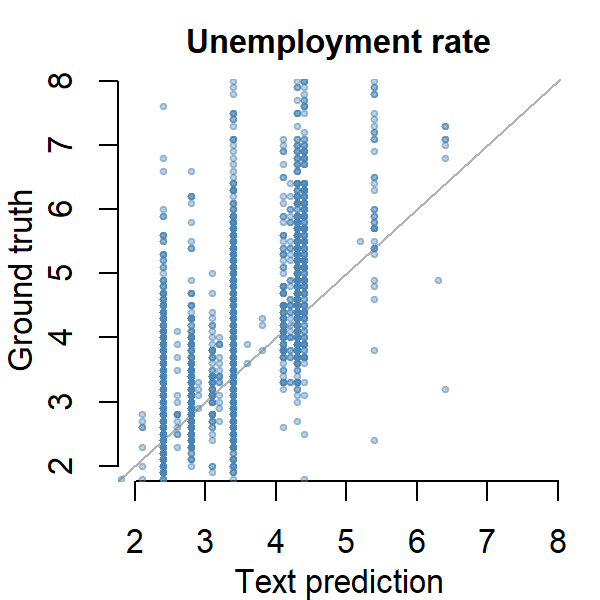}
\includegraphics[width=\ww\linewidth]{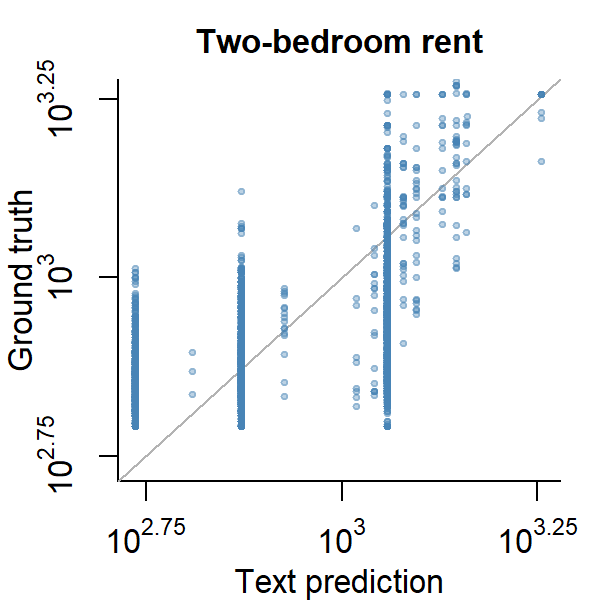}

\caption{Comparing actual values (ground truth) to the predicted values by LME (top row) and the text output (bottom row).}
\label{fig:scatters}
\end{figure}

Figure \ref{fig:scatters} plots the actual values against both the inferred values of LME (completion prompt, top row) and the text output (bottom row) on a few variables of the US counties dataset. The text output exactly recalls the population in a large number of counties, particularly for those counties with a large population. For the other variables, however, the text output is clustered on a few distinct values.\footnote{Clustering of values in the text outputs has also been observed by \citet{li2024can} in their analysis of LLMs' knowledge of regional statistics.}$^{,}$\footnote{Note that the values are not just clustered by US state. We typically observe several unique values within states as shown by Figure \ref{fig:n_unique} which depicts the number of unique ground-truth and text output values for each state.} By contrast, the distribution of LME predictions does not cluster but also does not reproduce the exact values. 

For the US states dataset, we also evaluate the performance of the text output and LME \textit{within} states. To obtain stable estimates we exclude states that have less than 50 counties. Figure \ref{fig:within_states} in the appendix compares the Spearman correlation between the ground truth and the text output and LME, respectively. Except for the population variable, we observe a performance advantage of LME over the text output in most states.\footnote{We cannot show the variable \textit{proportion of mortgages being delinquent} because it has a low coverage in the raw data, which is why we do not observe at least 50 non-missing observations in any state.}

\subsection{Robustness}
 We test whether we can improve the performance of LME by reducing the dimensionality of the embeddings using a PCA before training LME. For our two key datasets, US counties and US firms, Figures \ref{fig:performance_pca_US_spearman} and \ref{fig:performance_pca_yahoo_spearman} in the appendix respectively show that LME usually performs best without applying PCA. 

We also compare the performance by layer and see that across the key datasets, embeddings from the 25\textsuperscript{th} layer perform better on average than those from earlier or later layers (Figure \ref{fig:by_layer} in the appendix). However, the differences in performance are small with the exception of the lower performance when using layer 5, the earliest layer we examine.

\subsubsection{Comparing base models}

To test whether our finding that LME tends to beat the text output generalises across model families and model sizes, we run the cross-validation analysis on our two largest datasets (US counties and US listed firms) for all four LLMs.

\begin{figure}[!htb]
\begin{center}
\renewcommand\ww{.32}
\includegraphics[width=\ww\linewidth]{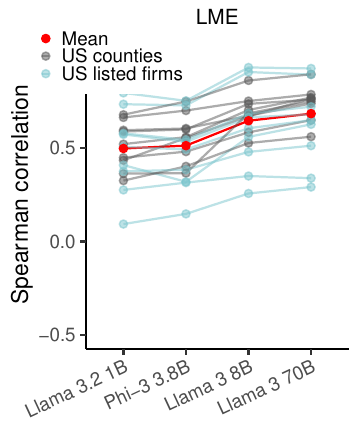}    
\includegraphics[width=\ww\linewidth]{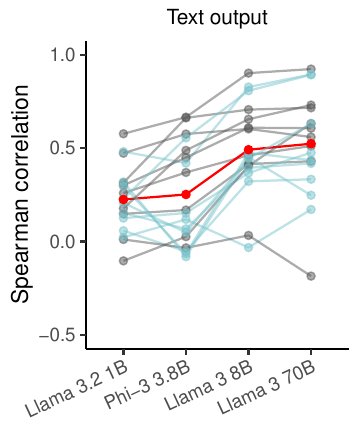}    
\includegraphics[width=\ww\linewidth]{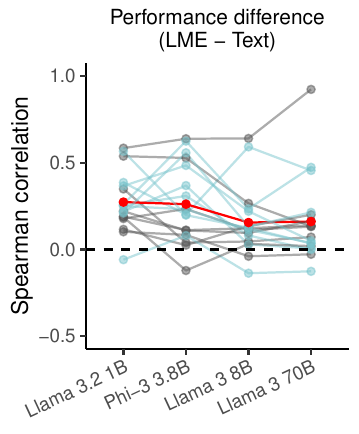}    
\caption{Performance of LLMs of different sizes.}
\label{fig:model_size}
\end{center}
\end{figure}

Figure \ref{fig:model_size} (left panel) shows how the performance of LME improves consistently with increasing model size.\footnote{As the models differ in their number of layers, we train LME on the embeddings of the last layer of each model. Furthermore, while LME on the embeddings of the Llama models work best when no dimensionality reduction is applied, Phi generally performs better when training LME on 25 PCA components. Thus, we applied the dimensionality reduction for Phi.}
The middle panel of the figure shows that the accuracy of the text output also increases with larger models, but less consistently. The right panel shows the difference in Spearman correlation of the two approaches (LME - text output). While the mean difference (in red) is larger for the small models, it is still large -- on average 16 percentage points -- for the 8B and 70B model.

\subsubsection{LME vs.\ a reasoning model} \label{sec:reasoning}

Since 2024, it has emerged that the performance of language models does not only scale with training-time compute (i.e.\ larger models and more training data) but also test-time compute. Allowing an LLM more time to generate responses, break down tasks into multiple steps, and reflect on its reasoning process significantly enhances its performance on reasoning tasks. We test whether this reasoning paradigm is also useful for our specific task: Can an LLM accurately reason about its internal knowledge of regions and firms, integrate this information, and ultimately provide precise estimates of relevant statistics?

We use the open-source reasoning model Qwen QwQ-32B \citep{qwq32b}. It is based on Qwen 2.5 32B \citep{qwen2.5}, an LLM not tuned to reason using test-time compute. By comparing these two models, we can directly measure the effect of the reasoning paradigm on accuracy.\footnote{In contrast to the previous experiments where we use a completion prompt, we use the question-answering prompt (see Section \ref{appx:prompting}) for the two Qwen models in order to elicit reasoning behaviour.} We pit the text output of the two models against LME trained on the embeddings of the final token of the prompt of the Qwen base model (Qwen 2.5 32B). Due to the high computational costs of the reasoning model, we only test it on a subsample of 500 entities per variable.

A closer look at the reasoning process reveals that the LLM tries to reconcile different information to provide the best estimates based on its knowledge.\footnote{For example, when asked for the proportion of delinquent mortgages in Atlantic County, New Jersey in 2019 one of many points the reasoning model made was ``Wait, I think that in 2019, the national delinquency rate was around 3-4\%, but that's the national average. Atlantic County might be different, especially if it's a coastal area that might have been affected by Hurricane Sandy in 2012, but that's a few years prior. Maybe the recovery from that could affect 2019 numbers? Not sure.'' Another sentence from the same answer: ``Hmm, I'm not entirely sure, but I think the national average was around 1.8\%, and New Jersey might be a bit higher. Since Atlantic County is a county with some economic challenges, maybe it's a bit higher than the state average."} However, despite the efforts, we do not observe a consistent increase in performance.

\begin{figure}[!htb]
\renewcommand\ww{.49}
\includegraphics[width=\ww\linewidth]{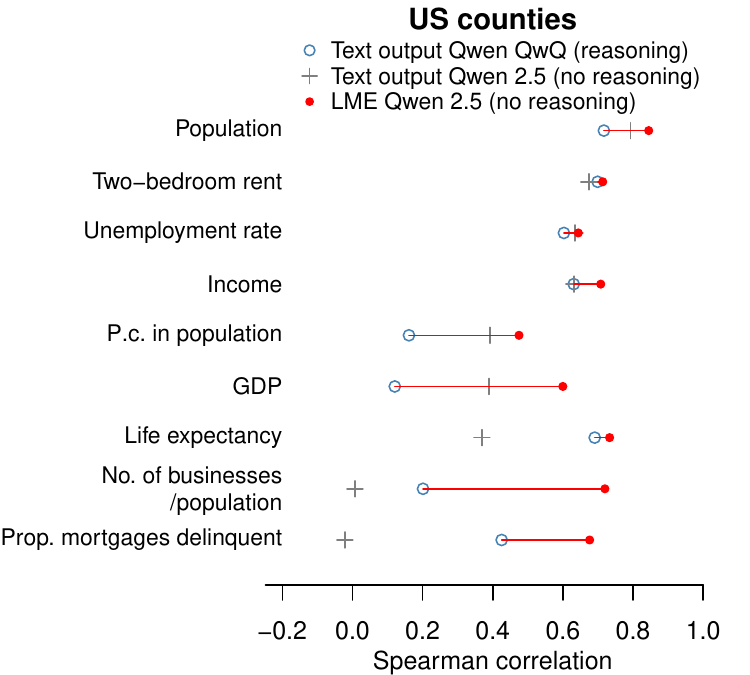}    
\includegraphics[width=\ww\linewidth]{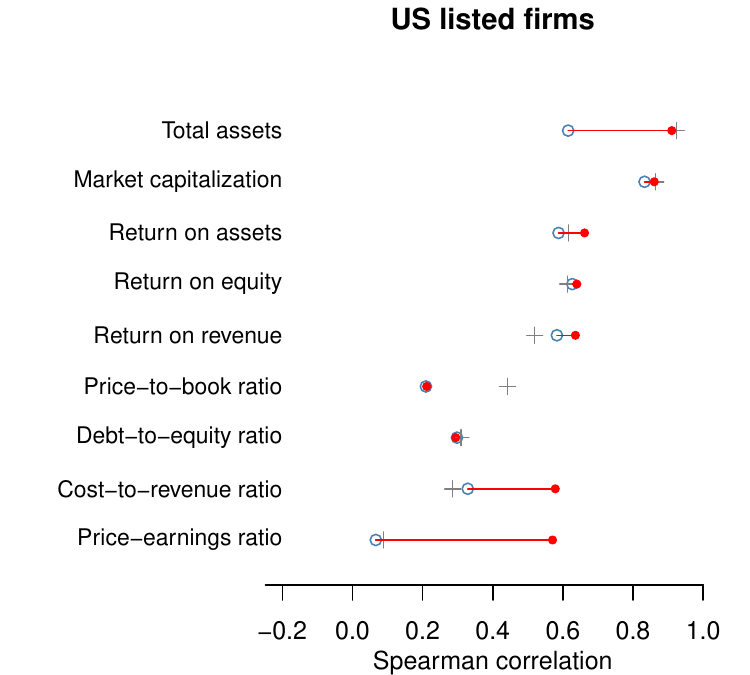}    
\caption{Cross-validation performance comparing reasoning model Qwen QwQ (text output) to Qwen 2.5 (text output + LME).}
\label{fig:reasoning}
\end{figure}

Figure \ref{fig:reasoning} shows the results for our two main datasets. Overall, we do not see a consistent improvement in performance when using the reasoning model to generate the text output. However, the reasoning model performs substantially better on the three US county variables on which the non-reasoning model performs worst. LME beats the text output of both models on most variables.
The difference in computational costs is significant. The reasoning model uses a median of 1767 output tokens, whereas Qwen 2.5 answers the question with a median of 8 tokens.

\subsection{Learning curves} \label{sec:learning}


\begin{figure}[!htb]
\renewcommand\ww{.325}
\includegraphics[width=\ww\linewidth]{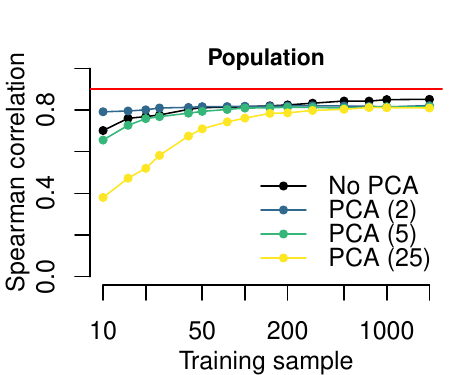}    
\includegraphics[width=\ww\linewidth]{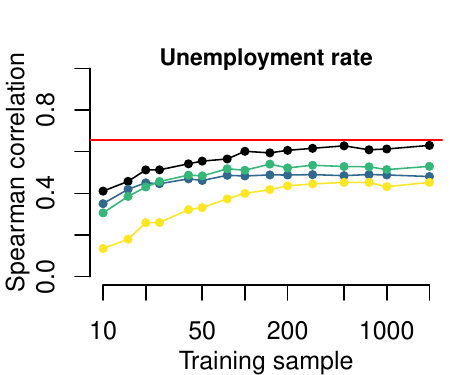}    
\includegraphics[width=\ww\linewidth]{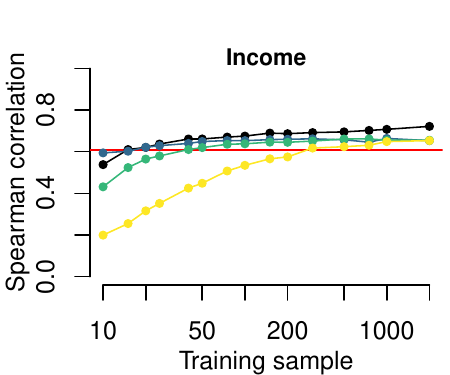}    \\
\includegraphics[width=\ww\linewidth]{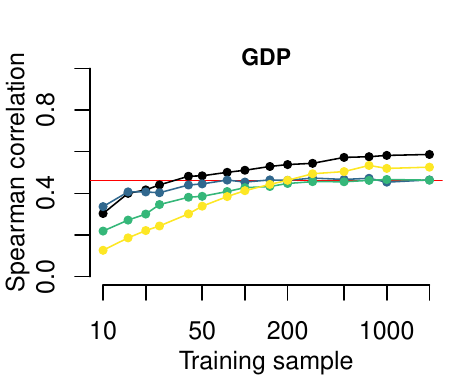}    
\includegraphics[width=\ww\linewidth]{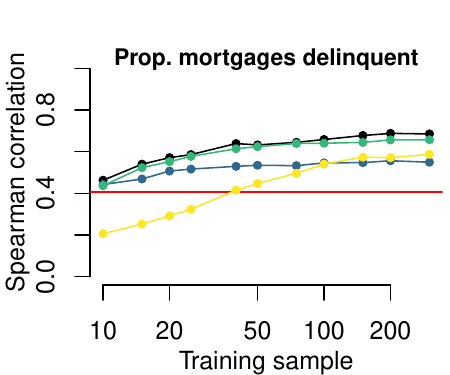}    
\includegraphics[width=\ww\linewidth]{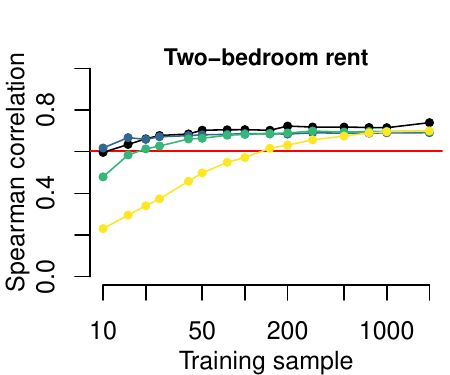}    

\vspace{.5cm}

\includegraphics[width=\ww\linewidth]{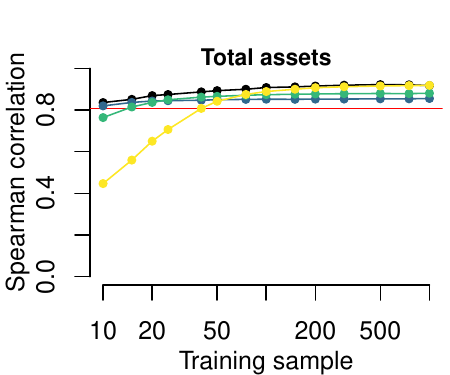}    
\includegraphics[width=\ww\linewidth]{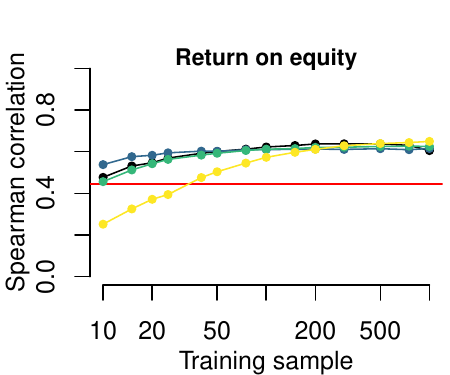}    
\includegraphics[width=\ww\linewidth]{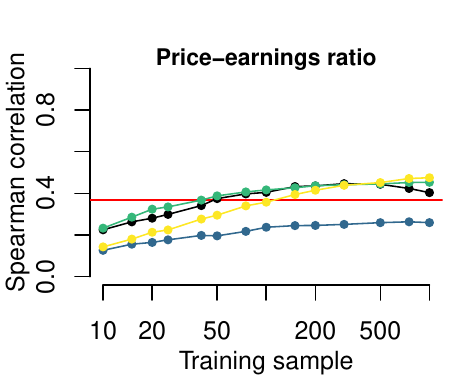}    
\caption{Learning curve analysis.}
\label{fig:learning}
\end{figure}

\begin{figure}[!htb]
\begin{center}
\includegraphics[width=.32\linewidth]{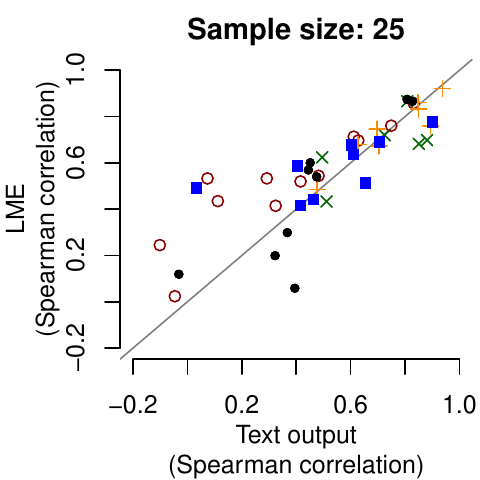}    
\includegraphics[width=.32\linewidth]{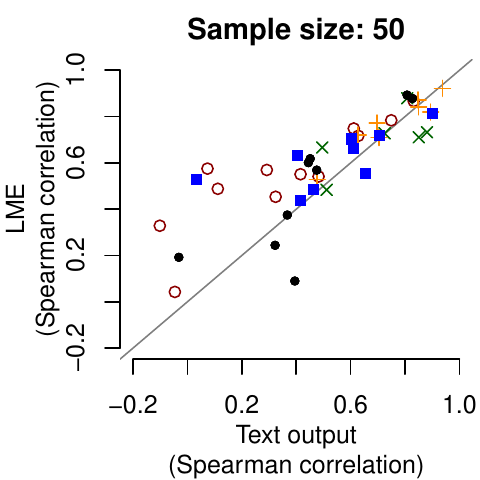}    
\includegraphics[width=.32\linewidth]{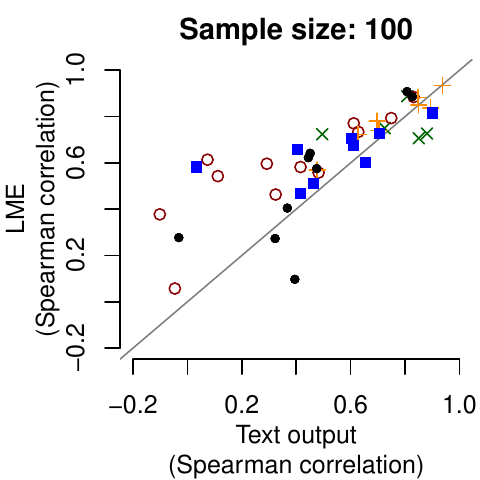}    
\includegraphics[width=.8\linewidth]{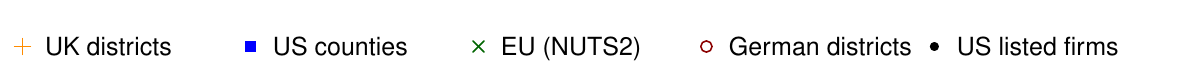}    
\end{center}
\caption{Comparison between text output and baseline LME with different training samples.}
\label{fig:learn_scatter}
\end{figure}

The cross-validation results show that we can infer statistical variables more accurately from the embeddings than from the text output. However, the 5-fold cross-validation setting assumes that 80\% of the values on the variable we are predicting are available to the modeller. If LME can reach a high accuracy with only a small number of labelled data points, its relevance for data processing tasks will be much higher. To test this, we conduct a learning curve analysis, estimating model performance as a function of the number of training samples. Figure \ref{fig:learning} compares the text output (red line) with LME trained on progressively larger training sets from the US counties and US firms datasets. LME's performance often converges with only a small number of training samples. Dimensionality reduction using PCA improves performance on some variables for very small sample sizes and when restricted to two components. Figure \ref{fig:learn_scatter} summarises the learning-curve results across all datasets, comparing the text output with the baseline LME at training sample sizes of 25, 50, and 100 observations. With only 25 samples, LME outperforms the text output for most variables, showing a mean Spearman correlation 5 percentage points higher.

\subsection{Transfer learning}  \label{sec:transfer_learning}
\subsubsection{Learning from embeddings of other variables} \label{sec:simple_transfer_learning}
The previous analyses assume that we have at least some labelled instances of a variable to estimate its values on other instances using the embeddings. Here, we test whether we can reliably estimate a regional or firm variable without access to any values of this variable. If the embeddings encode numbers in a consistent way across different variables, we should be able to predict variable $v_{test}$ using a model learned from other variables. More formally, given all variables $\mathbb{V}$ of a dataset with $n$ entities, we train a regression model on embeddings $E_i$ and outcome values $y_i$ of entities $i$, and test the models on variable $v_{test}$. Training observations are described as $\{(y_{vi}, E_{vi}): v \in  \mathbb{V} \setminus \{v_{test}\}, i = 1,...,n\}$ and test observations as $\{(y_{vi}, E_{vi}): v = v_{test}, i = 1,...,n\}$. Note that $E_v \neq E_w, \forall v \neq w$, as we use the completion prompt to obtain the embeddings, which also contains the name of the variables.

To ensure our model is flexible enough to learn generalised embeddings, we train a neural network with two layers (128 neurons and 32 neurons, Leaky ReLU activation function) as our transfer learning model. We use the Adam optimiser with an initial learning rate of $10^{-5}$. To avoid overfitting, we only learn for 20 epochs and apply a strong dropout of 0.5 on the first layer. We test this approach on all five datasets. All variables are transformed as described in Section \ref{sec:data} before they are normalised (mean of zero, standard deviation of one).

\begin{figure}[!htb]
\begin{center}
\renewcommand\ww{.49}
\includegraphics[width=\ww\linewidth]{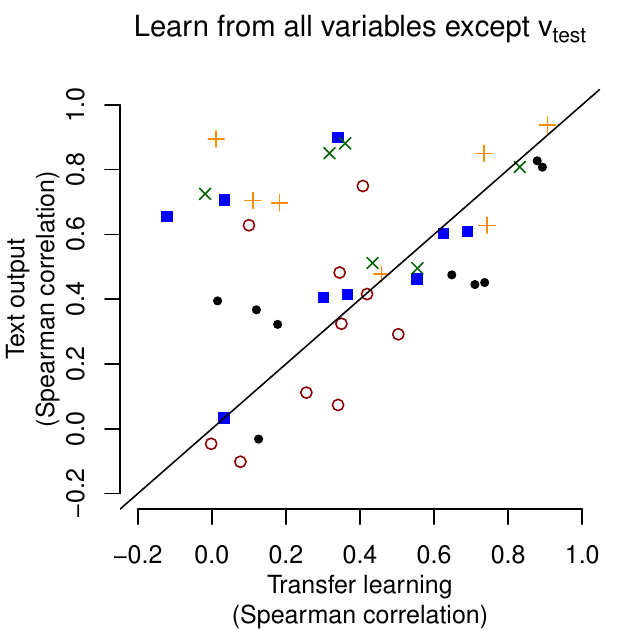}    
\includegraphics[width=\ww\linewidth]{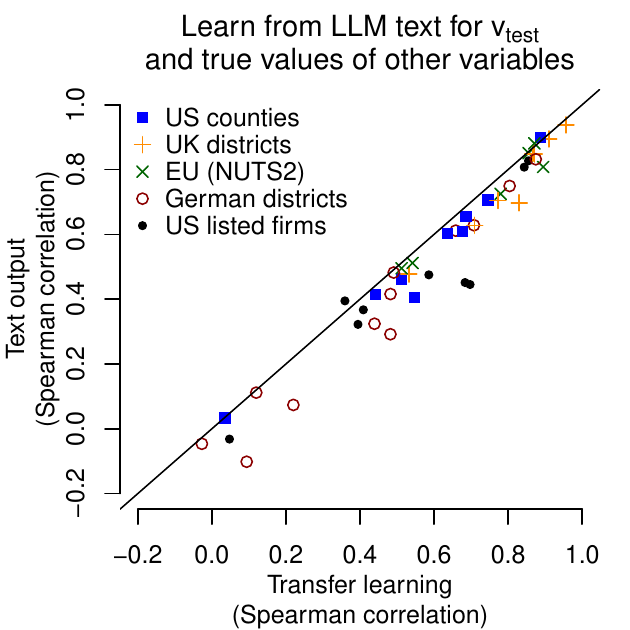}    
\caption{Transfer learning analysis.}
\label{fig:transfer_main}
\end{center}
\end{figure}

The left panel of Figure \ref{fig:transfer_main} compares the performance (Spearman correlation) of this approach to the text output. Each dot reflects one variable, for which a separate model was trained on all other variables of the same dataset. This approach to transfer learning does not beat the text output. While transfer learning is superior in some cases, it falls behind the text output most of the time. We obtain similar results when using a linear model instead of the neural network. We also test another simple transfer learning approach, where we predict variable $y$ in Dataset $A$ from the same variable of different datasets. For example, we train a model on the unemployment rate of the UK and Europe to predict the unemployment rate in the US. However, this transfer learning approach is also inferior to the text output on average, and sometimes by a large margin. 

Collectively, these transfer learning results show that embeddings do not generalise reliably to other variables. 

\subsubsection{Learning from text output}

Given the poor performance of the simple transfer learning approaches, we refine our method to exploit information about the variable we predict ($v_{test}$), even without access to ground truth values. Concretely, we use the text outputs of the LLM as noisy or biased labels of the variable of interest in the training set ($\overset{txt}{y}_{v_{test}}$). For the remaining variables we use the ground truth labels, as above. Formally, let 
$y^\dagger_{vi} \;:=\;
\begin{cases}
y_{vi}, & v \neq v_{test}\\
\overset{txt}{y}_{vi}, & v = v_{test}
\end{cases}$ and the training set be $\{\,  (y^\dagger_{vi}\, E_{vi}) \;:\; v \in \mathbb{V},\ i=1,\ldots,n_v \,\}$. The test set,  as before, is denoted as $\{(y_{vi}, E_{vi}): v = v_{test},i=1,...,n\}$. Even though the text labels are often inaccurate, they might provide enough information for the model to learn from the embeddings of the specific prompt of $v_{test}$. 

We use the same neural network set-up as above. The right panel of Figure \ref{fig:transfer_main} compares this performance of this transfer learning approach to the text output. This transfer learning approach almost always outperforms the text output even when the text output itself is accurate, on average by 7.2 percentage points.  This is an important finding. We can extract knowledge about a specific economic or financial variable from LLM embeddings \textit{without any} labelled instances of that variable. The result is robust to the choice of the hyperparameters of the neural network and we observe qualitatively similar, but slightly inferior performance, when using a linear model instead of the neural network.


\section{Leveraging LLM embeddings for imputation and super-resolution} \label{sec:application}

\subsection{Imputation}

\renewcommand\ww{.4}
\begin{figure}[!bh]
\centering
\includegraphics[width=\ww\linewidth]{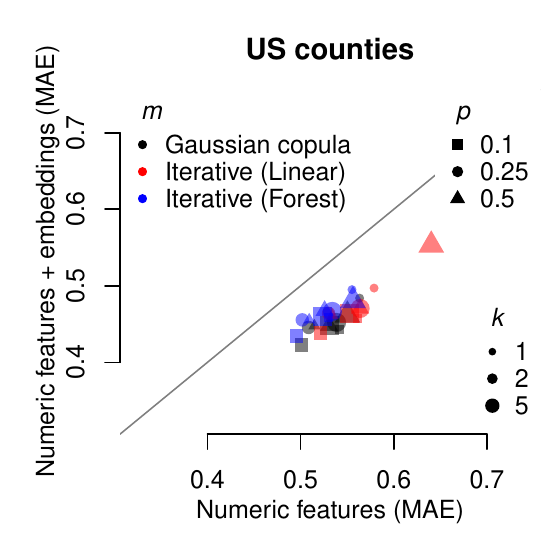}    
\includegraphics[width=\ww\linewidth]{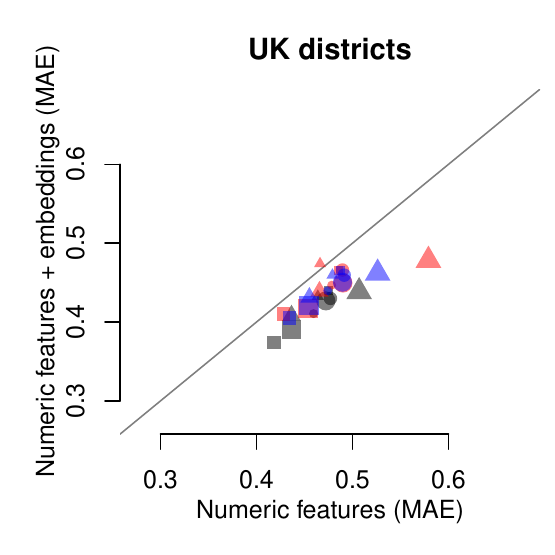}    
\includegraphics[width=\ww\linewidth]{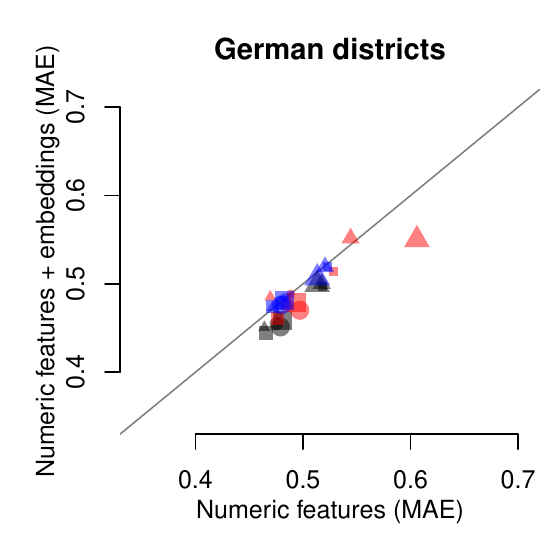}    
\includegraphics[width=\ww\linewidth]{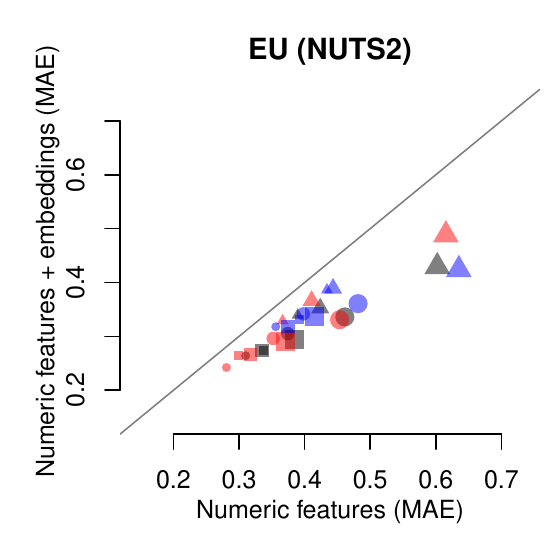}    
\includegraphics[width=\ww\linewidth]{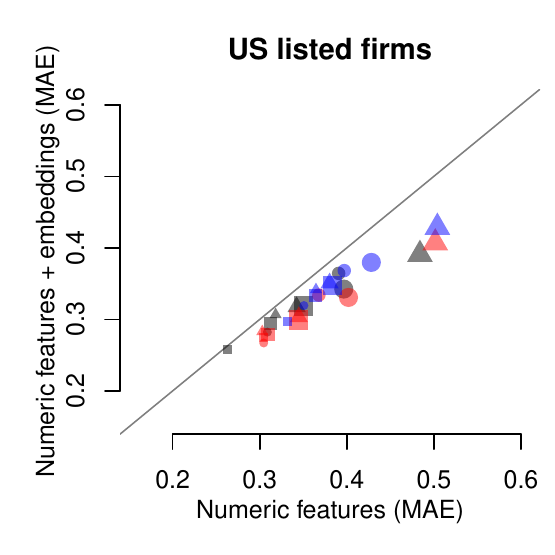}    
\caption{Imputation error when including or excluding the embedding vectors as features in the imputation method.}
\label{fig:impute_main}
\end{figure}

Our analyses have established that LLMs have rich knowledge about entities such as regions and firms. Here we test whether this knowledge can be exploited to improve the imputation of missing values, a common challenge in economic modelling. As in Section \ref{sec:results_main}, our approach does not assume the model has memorised the exact statistic to be imputed or that it can be recovered by prompting. Instead, we use the model’s embeddings as features, leveraging their generalised knowledge of entities. This makes the method applicable when the ground-truth statistic has not previously been produced or reported for certain entities and therefore can only be estimated.

To test whether embeddings can improve imputation accuracy, we compare two imputation strategies on our five datasets. The \textit{baseline} strategy estimates imputed values using only the $K$ numeric features (e.g.\ population and unemployment rate) in the dataset. The \textit{embedding} approach extends the feature matrix by appending the first $c$ PCA components of the embeddings.

To obtain embeddings for the entities, we use generic prompts~(see Section \ref{sec:setup}), which only contain the name of the entity (e.g.\ region or firm) and the year.
Although some of our datasets contain missing values, we do not impute them for evaluation because the true values are unknown, precluding accuracy assessment. Instead, we simulate missingness by randomly masking observed entries.

Our approach to introduce missing values is as follows: Given a dataset with $N$ observations and $K$ variables, all variables are transformed as described in Section \ref{sec:data} before they are normalised (mean of zero, standard deviation of 1). First, we sub-sample $k$ features, for each of which we randomly mask $p$\% of its values, creating the incomplete data $D_{miss}$. We then impute the missing values in $D_{miss}$ using imputation algorithm $m$.

We test three imputation methods $m$: using a low-rank Gaussian copula model \citep{zhao2020matrix} and iterative multivariate imputation using (i) Bayesian Ridge regression and (ii) random forest \citep{wilson2022miceforest}. We assess the accuracy of the imputation with the mean absolute error (MAE) between the ground truth and the reconstructed values.  

For each imputation method, we conduct a large array of experiments to ensure robustness of our results. Specifically, we consider all possible combinations of the parameter settings $k \in \{1,2,5\}, p \in \{0.1,0.25,0.5\}$. To obtain stable results, we replicate the experiments for each parameter combination 25 times, randomly choosing the instances and features which are imputed. As embedding features, we use the first $c=25$ PCA components of the 25\textsuperscript{th} layer of Llama 3 8B-Instruct. 

Figure \ref{fig:impute_main} presents the results across all five datasets and parameter settings $m$, $k$, and $p$. The embedding strategy consistently outperforms the baseline strategy. The improvement is particularly notable in some datasets. For instance, in the US counties dataset, the embedding strategy reduces the mean error by 14\% on average across parameter settings.

\subsection{Super-resolution} \label{sec:super_res}


\begin{table}[!tb]
\begin{center}
\begin{tabular}{llll}
\hline
& \multicolumn{2}{c}{Regional level} \\
Region & lower &  higher \\ 
\hline
US & 3142 counties & 51 states \\
EU (NUTS3) & 1165 NUTS-3 & 244 EU NUTS-2\\
Germany  & 401 districts & 38 gov.\ districts\\
\hline
\end{tabular}
\caption{Regional levels of super-resolution datasets}
\label{tab:superres_data}
\end{center}
\end{table}

We define super-resolution as the estimation of statistics at a lower geographical level using information available at a higher level when data for the lower level are unavailable or unreliable. Like imputation, this task has high practical relevance. Many economic indicators are published only at higher administrative levels, despite systematic sub-regional heterogeneity that we may wish to estimate. LME, exploiting an LLM's generalised knowledge about regions, could provide accurate estimates in this situation. For evaluation, however, lower-level ground truth is required, which is why we test our super-resolution approach only on data for which lower-level statistics are reported. 

We approach the super-resolution task by training LME on a higher regional level (e.g.\ unemployment rate of US states) to obtain estimates at the lower regional level (e.g.\ unemployment rate of US counties). We use the completion prompt for both geographical levels. We test this approach on three of our datasets (US, EU, Germany) for which we were able to find data on two regional levels, as shown in Table \ref{tab:superres_data}.

\renewcommand\ww{.49}

\begin{figure}[!htb]
\includegraphics[width=\ww\linewidth]{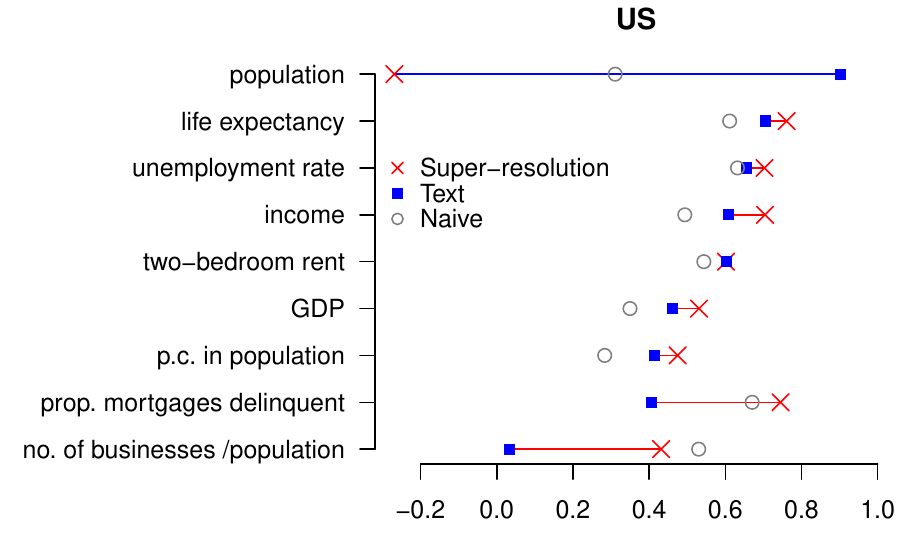}    
\includegraphics[width=\ww\linewidth]{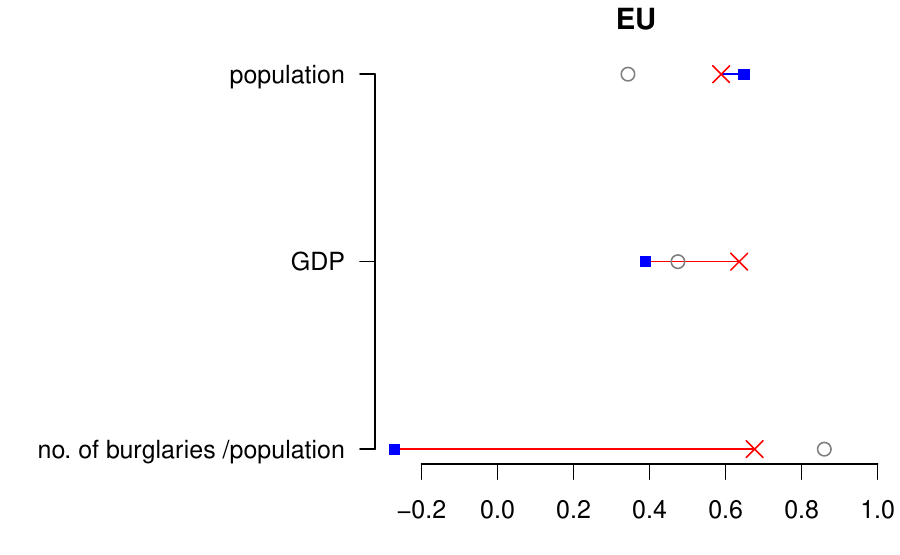}    
\includegraphics[width=\ww\linewidth]{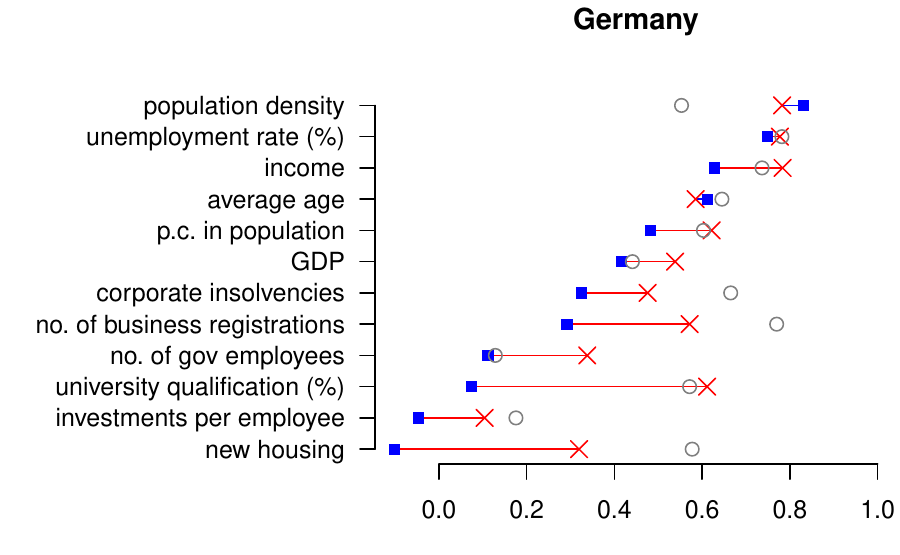} 
\caption{Super-resolution analysis.}
\label{fig:super-resolution}
\end{figure}

We compare this super-resolution approach with the text output at the lower level and with a \textit{naive} method, which projects higher-level region values onto the lower level. For example, under the naive method, the unemployment rate of all counties in Alabama is set equal to the unemployment rate reported for the state of Alabama. Figure \ref{fig:super-resolution} presents the results, ordering the variables by decreasing performance of the  text output.

The super-resolution approach performs well, surpassing the text output on most variables. A notable outlier is population, where the super-resolution performs worse than random. The naive approach performs well on many of the variables in Germany but falls behind the other two approaches on the US data. Given the small training sample size, in particular for the US and Germany (51 and 38 observations at the higher geographical level, respectively), the performance of the super-resolution approach is impressive. Super-resolution is a particular transfer learning approach and thus supports our previous finding (Section \ref{sec:transfer_learning}) that the embeddings can be used for accurate estimation outside the domain of the training set. 

\section{Conclusion} \label{sec:conclusion}

We have shown that LLMs know more than they say. A linear model trained on the LLM's hidden states of only a few dozen labelled samples often outperforms the LLM's text output in the estimation of statistics -- in particular when the target variable is a less common statistic. 
A key contribution of our work is the transfer learning approach which shows that we can estimate target variables more accurately than the text output without any labelled data on that target variable. Finally, we provide robust evidence that the hidden states can be used for data processing tasks. In particular, adding a low-rank representation of the hidden states to the feature matrix consistently improves the performance of standard imputation approaches in all of our datasets. Our super-resolution results show that, in most cases, we can beat the text output in estimating statistics on a lower geographical level by learning from the higher geographical level.

We considered several LLMs with 1 to 70 billion parameters and show that our result that LLMs know more than they say holds across all of them, including a reasoning model that uses extensive test-time compute. We leave it for future work to test whether this finding also holds for the largest and most accurate open-source models such as Llama 3 405B or DeepSeek V3 with 671 billion parameters.  In this study we exclusively worked with public data sources; it would be interesting to examine the extent to which our results hold for proprietary datasets, which are less likely to be observed in the training corpus of open-source LLMs. Finally, future research could investigate how useful the hidden states representing firms or geographic entities can be for other data processing tasks, such as outlier detection.

\clearpage


\bibliographystyle{chicago}
\bibliography{bib}

\pagebreak

\appendix

\section*{Appendix}

\renewcommand{\thefigure}{A\Roman{figure}}
\renewcommand{\thetable}{A\Roman{table}}
\setcounter{figure}{0}
\setcounter{table}{0}

\section{Datasets} \label{appx:data}

\subsection{US counties}

The data on US counties for the year 2019 is one of our two main datasets and we used a variety of sources to assemble it.
We obtained income and GDP data from the website of the Bureau of Economic Analysis\footnote{\url{https://www.bea.gov/data}}. The unemployment data was sourced from the Economic Research Service of the US Department of Agriculture\footnote{https://www.ers.usda.gov/data-products/county-level-data-sets}. Life expectancy data is provided by the Institute for Health Metrics and Evaluation in the Global Health Data Exchange database. We obtained mortgage delinquency rates from the Consumer Financial Protection Bureau.\footnote{\url{https://www.consumerfinance.gov/data-research/mortgage-performance-trends/download-the-data}} Data on the number of businesses was downloaded from the County Business Patterns Tables from the United States Census Bureau.\footnote{\url{https://www.census.gov/programs-surveys/cbp/data/tables.html}} Information on the median rent is available from the website of the Office of Policy Development and Research.\footnote{\url{https://www.huduser.gov/portal/datasets/50per.html}} The remaining variables we use come from the R package \texttt{usdata} \citep{usdata} which lists the US Census and the American Community Survey as sources.

\subsection{UK regions}

We gathered data for the year 2019 for the 374 local authority districts (LADs) in the United Kingdom.
Most variables were sourced from the Office for National Statistics (ONS) using API access via the R package \texttt{onsr} \citep{onsr}. Some series are not directly accessible via API and were downloaded from the official websites of the ONS, Northern Ireland Statistics and Research Agency (NISRA) and the Scottish and Welsh government websites.

\subsection{EU regions}

All the EU series for the year 2019 are collected from Eurostat, an official database of the European Union. We access the series at the NUTS2 and NUTS3 levels via API calls using the R package \texttt{eurostat} \citep{lahti2017,eurostat}. Note that fewer series are available on the NUTS3 level.

\subsection{German districts}
All series for the year 2019 are downloaded directly from the Regionalatlas Deutschland\footnote{\url{https://regionalatlas.statistikportal.de/}} by the Federal Statistical Office of Germany. 

\subsection{US listed firms}

We collected financial information as of the end of 2022\footnote{We could not access data from 2019, as we do for the other datasets, because the free API access is limited to the most recent years.}, including balance sheet, income and share price data, of US listed firms using the Yahoo Finance\footnote{\url{https://finance.yahoo.com/}} free API access. We excluded values on the variables \textit{return on revenue} and \textit{return on equity} when firms have negative values on stockholders' equity or total revenue, respectively.

\section{Methodology} \label{appx:methods}

\subsection{Computational requirements} \label{appx:computation_time}

We used an NVIDIA Tesla V100 with 16 GB of video memory to generate the LLM's embeddings and generate text. With 16-bit (fp16) parameters, this memory was sufficient for all models except Llama 70B variant. Using low-priority virtual machines, compute costs were under US \$0.80 per hour for the GPU. Our baseline Llama 3 8B produced embeddings for a query in ~0.1 seconds. We also tested the speed of a quantised version of Llama 3 8B running locally on an AMD Ryzen 7 3700X CPU. This required ~0.9 seconds per query.

\subsection{Prompting} \label{appx:prompting}

In addition to the baseline completion prompting strategy (see Section \ref{sec:setup}), we also tested three alternative strategies. First, we apply the standard prompting template that is provided via the \texttt{apply\_chat\_template} function in the Transformers package \citep{wolf2020transformers}, referred to as question-answer prompting in this paper. For example, \textit{What was the unemployment rate  in Orange County, California in 2019?} 
This led to a lower response rate for US listed firms, with the model occasionally refusing to give an answer. Second, we test 5-shot prompting, where we provide 5 randomly chosen Q\&A pairs to the model to learn from. Finally, we use chain-of-thought prompting, adding the phrase ``think carefully before giving an answer" to the system prompt. The system prompts for the different prompting strategies are shown in Table \ref{tab:sys_prompt}.

Table \ref{tab:response_rate} indicates the response rates and Figure \ref{fig:prompting} compares the text output's performance of different prompting strategies. On average, the completion strategy has the highest response rate and performs best.

\begin{table}[!tb]
\begin{center}
\begin{tabular}{lp{9cm}}
Prompt type & System prompt \\ \hline
Completion prompt & You are a helpful assistant. \\
Question-answer prompt & You are a helpful assistant. If you do not know the answer to the question, provide your best estimate. Answer shortly like this. 'My answer: \{number\}\\
5-shot prompt & You are a helpful assistant. Answer the question. Use the format provided in the examples. \\
Chain-of-thought & You are a helpful assistant. If you do not know the answer to the question, provide your best estimate. Think carefully before giving an answer. Once you have the answer just state it like this. 'My final answer: \{number\}'.  \\
\hline
\end{tabular}
\caption{System prompts}
\label{tab:sys_prompt}
\end{center}
\end{table}

\footnotesize
\begin{longtable}{p{6.5cm}lp{2cm}ll} 
  \hline
Variable & Completion & Chain-of-thought & Few-shot & Q\&A \\ 
  \hline
\textbf{US counties} & \\ \hline
GDP per capita (in \$) & \textbf{1.00} & \textbf{1.00} & 0.99 & \textbf{1.00} \\ 
life expectancy at birth & \textbf{1.00} & \textbf{1.00} & \textbf{1.00} & \textbf{1.00} \\ 
median monthly two-bedroom rent (in \$) & \textbf{1.00} & \textbf{1.00} & 0.98 & \textbf{1.00} \\ 
personal income per capita (in \$) & \textbf{1.00} & \textbf{1.00} & 0.99 & \textbf{1.00} \\ 
population & \textbf{1.00} & \textbf{1.00} & \textbf{1.00} & \textbf{1.00} \\ 
unemployment rate & \textbf{1.00} & 0.99 & 0.99 & \textbf{1.00} \\ 
number of business establishments per 100,000 people & \textbf{1.00} & \textbf{1.00} & \textbf{1.00} & \textbf{1.00} \\ 
percentage change in population compared to the previous year & \textbf{1.00} & \textbf{1.00} & 0.98 & \textbf{1.00} \\ 
proportion of mortgages being 90 or more days delinquent & \textbf{1.00} & 0.99 & 0.20 & \textbf{1.00} \\ 
\hline
\textbf{US listed firms} & \\ \hline
cost-to-revenue ratio & \textbf{1.00} & 0.97 & 0.69 & 0.83 \\ 
debt-to-equity ratio & \textbf{1.00} & 0.96 & 0.99 & 0.79 \\ 
market capitalization & \textbf{1.00} & 0.94 & 0.99 & 0.97 \\ 
price-earnings ratio & \textbf{1.00} & 0.88 & 0.99 & 0.91 \\ 
price-to-book ratio & \textbf{1.00} & 0.95 & 0.99 & 0.92 \\ 
return on assets & \textbf{1.00} & 0.88 & 0.99 & 0.96 \\ 
return on equity & \textbf{1.00} & 0.92 & 0.82 & 0.88 \\ 
return on revenue & \textbf{1.00} & 0.98 & 0.84 & 0.91 \\ 
total assets & \textbf{1.00} & 0.98 & 0.99 & 0.96 \\ 
\hline
\caption{Response rate when using different prompting strategies}
\label{tab:response_rate}
\end{longtable}

\renewcommand\ww{.7}

\begin{figure}[h]
\begin{center}
\includegraphics[width=\ww\linewidth]{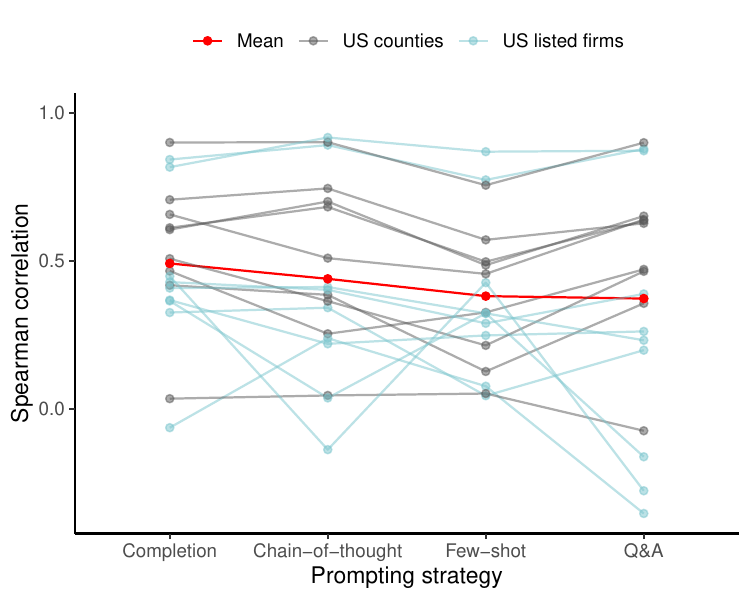}
\caption{Performance of text output when using different prompting strategies.}
\label{fig:prompting}
\end{center}
\end{figure}

\normalsize

\subsection{Extraction of numeric estimates from text output} \label{appx:number_extract}
Extracting numeric values from text outputs of smaller, open-source LLMs is challenging due to inconsistencies in the response structure and the numerical representation. We develop a rich extraction function with regular expressions. Firstly, we remove date strings (as these also contain numbers) and format numeric values by removing digit group separators before we convert answers into consistent units for the different variables. Additionally, with Q\&A prompting, LLMs typically provide the answer first and then offer an explanation. In contrast, for other strategies, LLMs tend to explain first and return the answer at the end. Therefore, based on the prompting strategy, we apply an additional conditional statement to select either the first or the last numeric value as the final prediction. 
\clearpage

\section{Additional results}

\begingroup
\addtolength{\tabcolsep}{-2pt}

\begingroup
\addtolength{\tabcolsep}{-2pt}

\footnotesize
\begin{longtable}{p{6.5cm}llr|ll} 
    \hline
    && \multicolumn{2}{c}{\textsc{Spearman}} &  \multicolumn{2}{c}{\textsc{Pearson}} \\
   Variable & Transfor-& LME & Text & LME & Text \\
   & mation& &&& \\

    \hline
    \endfirsthead
    
    \hline
    && \multicolumn{2}{c}{\textsc{Spearman}} &  \multicolumn{2}{c}{\textsc{Pearson}} \\
   Variable & Transfor-& LME & Text & LME & Text \\
   & mation& &&& \\    \hline
    \endhead
    
    \hline
    \multicolumn{6}{r}{\textit{Continued on next page}} \\ 
    \endfoot
    
    \endlastfoot

  \hline
\textbf{US firms}  & &&&& \\  
  return on assets & cubic & \textbf{0.69} (0.69-0.70) & 0.45 & \textbf{0.76} (0.75-0.76) & 0.55 \\ 
  return on revenue & cubic & \textbf{0.60} (0.60-0.61) & 0.48 & \textbf{0.66} (0.66-0.67) & 0.54 \\ 
  market capitalization & log & \textbf{0.91} (0.91-0.91) & 0.83 & \textbf{0.91} (0.91-0.91) & 0.50 \\ 
  return on equity & cubic & \textbf{0.67} (0.66-0.67) & 0.44 & \textbf{0.68} (0.68-0.68) & 0.50 \\ 
  total assets & log & \textbf{0.94} (0.93-0.94) & 0.81 & \textbf{0.92} (0.92-0.92) & 0.48 \\ 
  price-earnings ratio & cubic & \textbf{0.50} (0.49-0.51) & 0.37 & \textbf{0.46} (0.45-0.47) & 0.21 \\ 
  price-to-book ratio & cubic & 0.28 (0.26-0.29) & \textbf{0.39} & 0.14 (0.11-0.15) & \textbf{0.12} \\ 
  debt-to-equity ratio & cubic & \textbf{0.37} (0.36-0.38) & 0.32 & \textbf{0.18} (0.15-0.19) & 0.08 \\ 
  cost-to-revenue ratio & log & \textbf{0.60} (0.59-0.62) & -0.03 & \textbf{0.48} (0.46-0.49) & 0.02 \\ 
   \hline
\textbf{US counties}  & &&&& \\  
population & log & 0.87 (0.86-0.87) & \textbf{0.90} & 0.89 (0.88-0.89) & \textbf{0.91} \\ 
  life expectancy at birth & no & \textbf{0.75} (0.74-0.76) & 0.71 & \textbf{0.74} (0.73-0.74) & 0.69 \\ 
  unemployment rate & no & 0.63 (0.60-0.67) & \textbf{0.65} & 0.63 (0.60-0.67) & \textbf{0.64} \\ 
  median monthly two-bedroom rent (in \$) & log & \textbf{0.75} (0.74-0.75) & 0.60 & \textbf{0.85} (0.84-0.85) & 0.63 \\ 
  personal income per capita (in \$) & no & \textbf{0.73} (0.70-0.73) & 0.61 & \textbf{0.72} (0.71-0.73) & 0.63 \\ 
  GDP per capita (in \$) & log & \textbf{0.60} (0.59-0.62) & 0.46 & \textbf{0.56} (0.53-0.57) & 0.42 \\ 
  proportion of mortgages being 90 or more days delinquent & no & \textbf{0.70} (0.66-0.72) & 0.41 & \textbf{0.69} (0.65-0.71) & 0.41 \\ 
  percentage change in population compared to the previous year & no & \textbf{0.51} (0.50-0.52) & 0.41 & \textbf{0.48} (0.47-0.49) & 0.38 \\ 
  number of business establishments per 100,000 people & log & \textbf{0.69} (0.68-0.70) & 0.03 & \textbf{0.68} (0.67-0.69) & 0.04 \\ 
   \hline
\textbf{UK districts}  & &&&& \\  
average house price (in £) & no & \textbf{0.94} (0.91-0.96) & \textbf{0.94} & \textbf{0.89} (0.86-0.92) & \textbf{0.93} \\ 
  population & log & 0.85 (0.84-0.86) & \textbf{0.89} & 0.85 (0.81-0.86) & \textbf{0.91} \\ 
  median age & no & \textbf{0.89} (0.88-0.91) & 0.85 & \textbf{0.89} (0.88-0.91) & 0.82 \\ 
  percentage of the population that live in income deprivation & no & \textbf{0.87} (0.85-0.88) & 0.85 & \textbf{0.86} (0.84-0.88) & 0.81 \\ 
  median annual gross pay (in £) & no & \textbf{0.77} (0.76-0.80) & 0.63 & \textbf{0.83} (0.81-0.84) & 0.72 \\ 
  life expectancy for females at birth & no & \textbf{0.80} (0.76-0.82) & 0.70 & \textbf{0.76} (0.70-0.81) & 0.70 \\ 
  unemployment rate & no & \textbf{0.75} (0.70-0.78) & 0.70 & \textbf{0.73} (0.67-0.76) & 0.68 \\ 
  GDP per capita (in £) & log & \textbf{0.63} (0.59-0.69) & 0.48 & \textbf{0.64} (0.58-0.69) & 0.66 \\ 
   \hline
\textbf{German districts}  & &&&& \\  
number of hospital beds per 1,000 inhabitants & no & \textbf{0.53} (0.51-0.55) &  & \textbf{0.52} (0.49-0.55) &  0.00\footnote{With constant responses across districts, the sample correlation is not defined. We record performance as 0.}\\ 
  population density (per square km) & log & \textbf{0.90} (0.88-0.90) & 0.83 & \textbf{0.91} (0.90-0.92) & 0.82 \\ 
  unemployment rate (\%) & no & \textbf{0.83} (0.82-0.84) & 0.75 & \textbf{0.81} (0.79-0.83) & 0.74 \\ 
  disposable income per capita (in \euro) & no & \textbf{0.77} (0.76-0.78) & 0.63 & \textbf{0.76} (0.74-0.78) & 0.60 \\ 
  average age of the population & no & \textbf{0.82} (0.81-0.84) & 0.61 & \textbf{0.84} (0.83-0.85) & 0.57 \\ 
  GDP per capita (in \euro) & log & \textbf{0.69} (0.66-0.70) & 0.42 & \textbf{0.73} (0.70-0.74) & 0.43 \\ 
  percentage change in population compared to the previous year & no & \textbf{0.57} (0.54-0.60) & 0.48 & \textbf{0.57} (0.52-0.61) & 0.42 \\ 
  number of business registrations per 10,000 inhabitants & no & \textbf{0.63} (0.57-0.66) & 0.29 & \textbf{0.66} (0.59-0.69) & 0.36 \\ 
  number of filed corporate insolvencies per 10,000 taxable companies & no & \textbf{0.49} (0.45-0.52) & 0.32 & \textbf{0.59} (0.56-0.62) & 0.34 \\ 
  percentage of school leavers with a general university entrance qualification (\%) & no & \textbf{0.65} (0.63-0.67) & 0.07 & \textbf{0.63} (0.60-0.65) & 0.13 \\ 
  number of government employees per 1,000 inhabitants & no & \textbf{0.63} (0.62-0.65) & 0.11 & \textbf{0.63} (0.61-0.65) & 0.08 \\ 
  number of completed housing units per 1,000 inhabitants & no & \textbf{0.47} (0.41-0.52) & -0.10 & \textbf{0.42} (0.35-0.46) & 0.04 \\ 
  total investments per employee (in \euro) & no & \textbf{0.12} (0.04-0.18) & -0.05 & \textbf{0.07} (-0.01-0.13) & -0.09 \\ 
   \hline
\textbf{EU (NUTS2)}  & &&&& \\  
life expectancy at birth & no & 0.70 (0.54-0.79) & \textbf{0.85} & 0.86 (0.82-0.89) & \textbf{0.91} \\ 
  unemployment rate & no & 0.75 (0.52-0.82) & \textbf{0.88} & 0.78 (0.48-0.85) & \textbf{0.87} \\ 
  net disposable household income (in \euro) & no & \textbf{0.92} (0.89-0.94) & 0.81 & \textbf{0.94} (0.91-0.95) & 0.82 \\ 
  population & log & \textbf{0.73} (0.69-0.77) & 0.72 & \textbf{0.75} (0.71-0.81) & 0.68 \\ 
  risk-of-poverty rate & no & \textbf{0.55} (0.35-0.61) & 0.51 & \textbf{0.47} (0.34-0.59) & 0.50 \\ 
  GDP per capita (in \euro) & log & \textbf{0.75} (0.65-0.79) & 0.50 & \textbf{0.76} (0.67-0.80) & 0.46 \\ 
  number of burglaries per hundred thousand inhabitants & no & \textbf{0.55} (0.27-0.62) & -0.43 & \textbf{0.47} (0.28-0.56) & -0.01 \\ 
   \hline
\caption{Cross-validation performance of the baseline LLM comparing LME with text output. For each performance metric, the best-performing approach is shown in bold.
LME performance is based on 25 iterations of 5-fold cross-validation. Numbers before the parentheses indicate the mean; numbers in parentheses show the minimum and maximum performance across the 25 iterations.}
\label{tab:tab_cv_all}
\end{longtable}
\addtolength{\tabcolsep}{2pt}
\endgroup

\renewcommand\ww{.49}

\begin{figure}
 \includegraphics[width=\ww\linewidth]{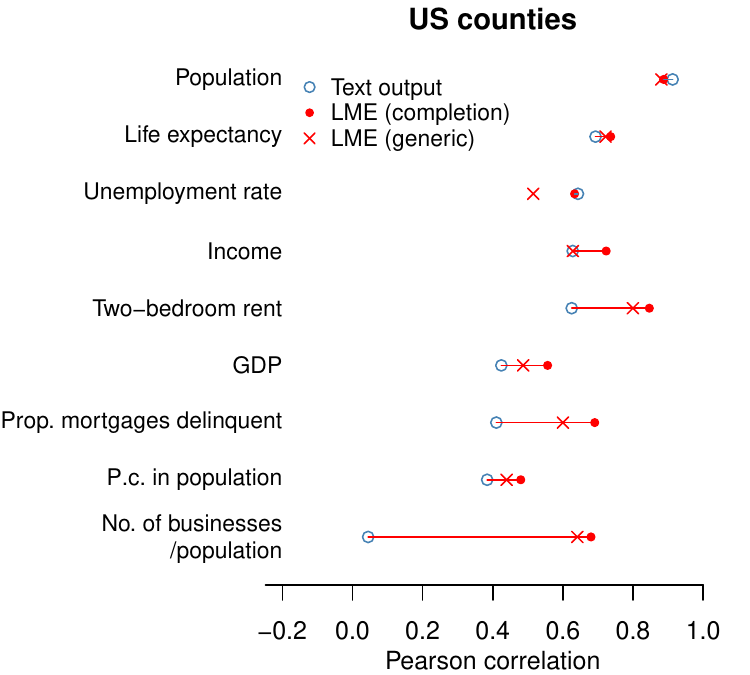} 
 \includegraphics[width=\ww\linewidth]{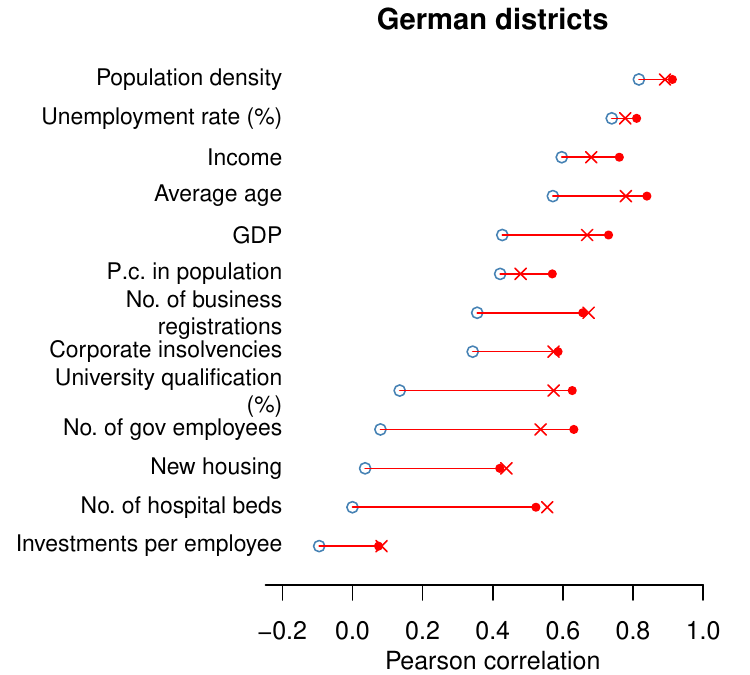} \\
 \includegraphics[width=\ww\linewidth]{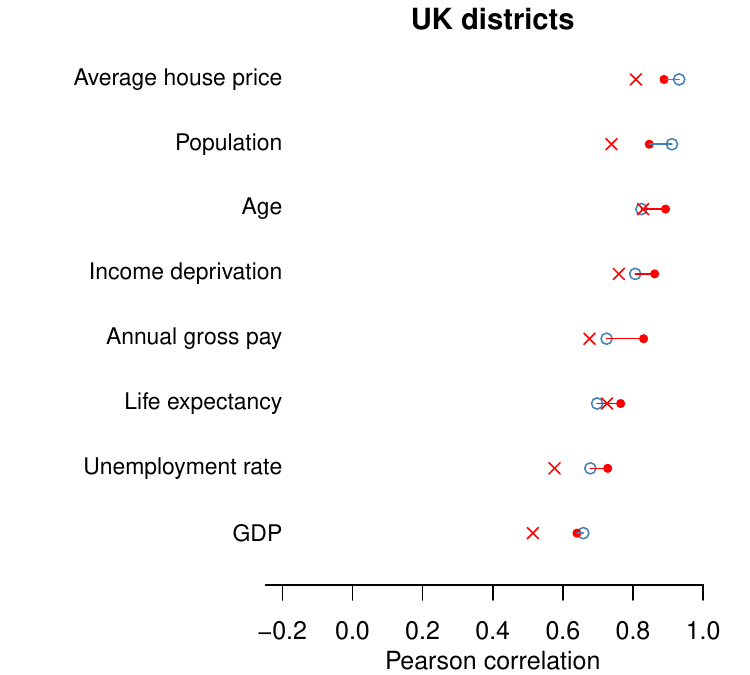}
  \includegraphics[width=\ww\linewidth]{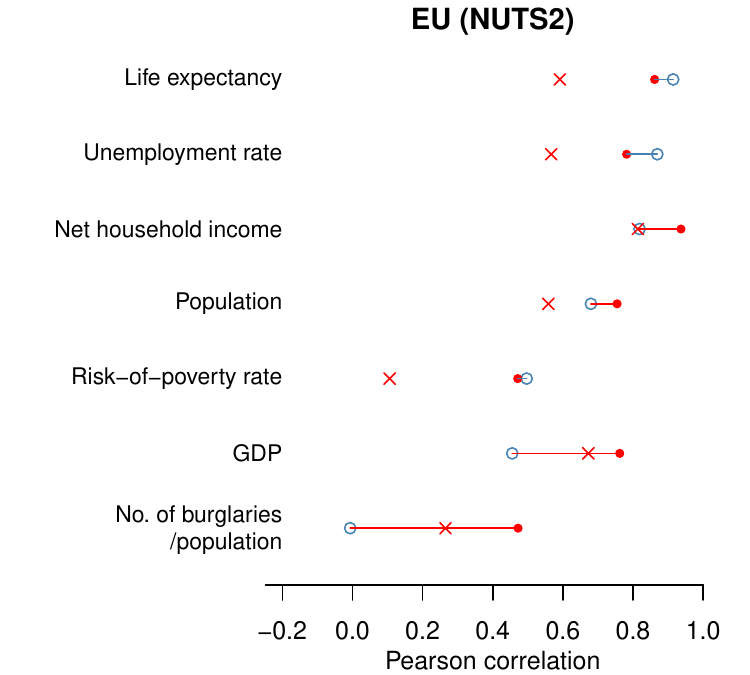}
  \includegraphics[width=\ww\linewidth]{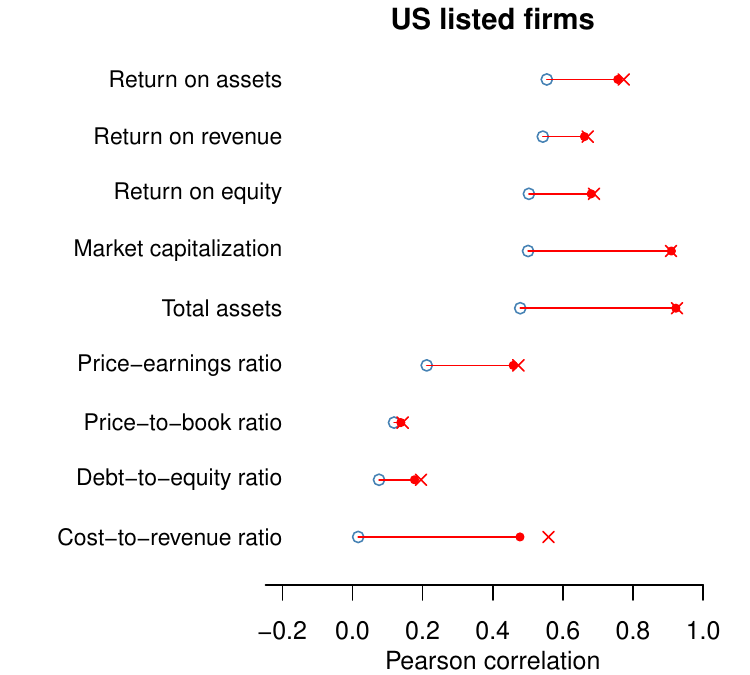}
\caption{Cross-validation performance using Pearson correlation as the performance metric.}
\label{fig:bars_pearson}
\end{figure}

\renewcommand\ww{.32}
\begin{figure}
\includegraphics[width=\ww\linewidth]{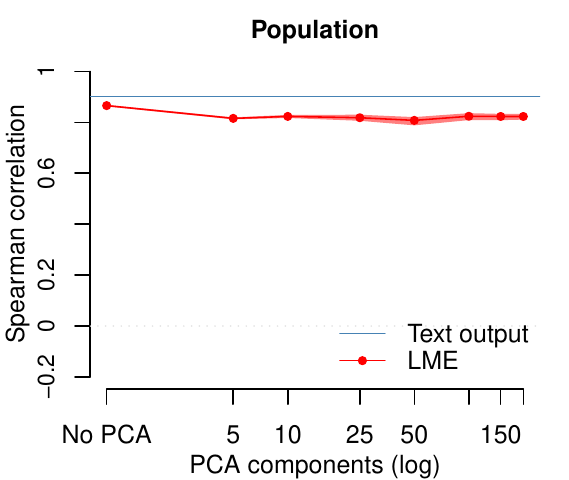}
\includegraphics[width=\ww\linewidth]{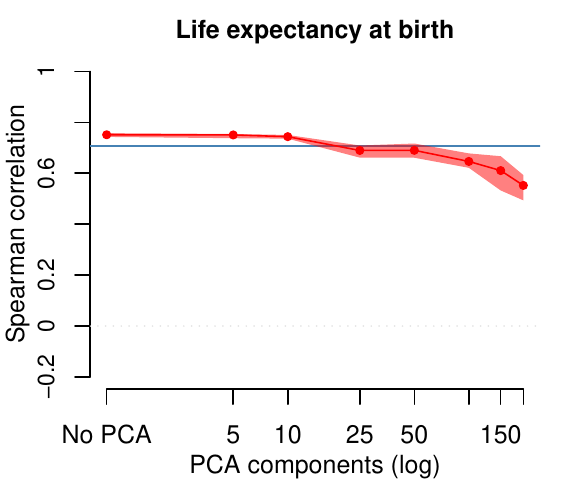}
\includegraphics[width=\ww\linewidth]{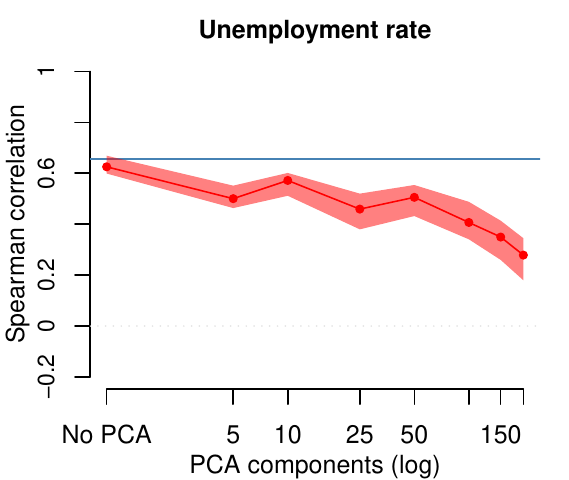}

\includegraphics[width=\ww\linewidth]{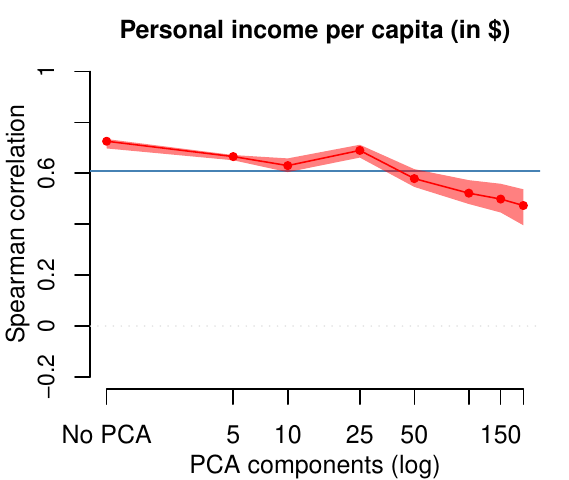}
\includegraphics[width=\ww\linewidth]{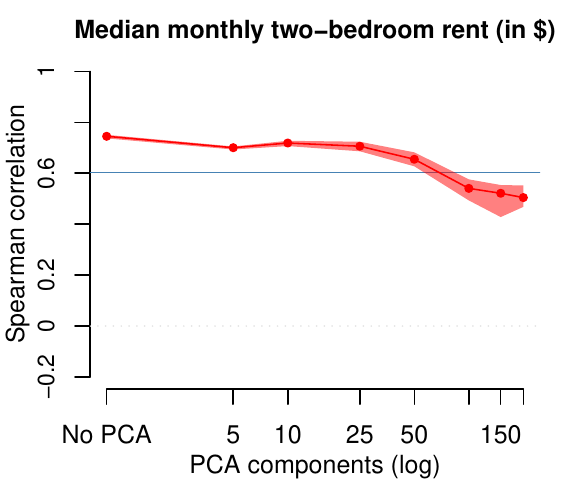}
\includegraphics[width=\ww\linewidth]{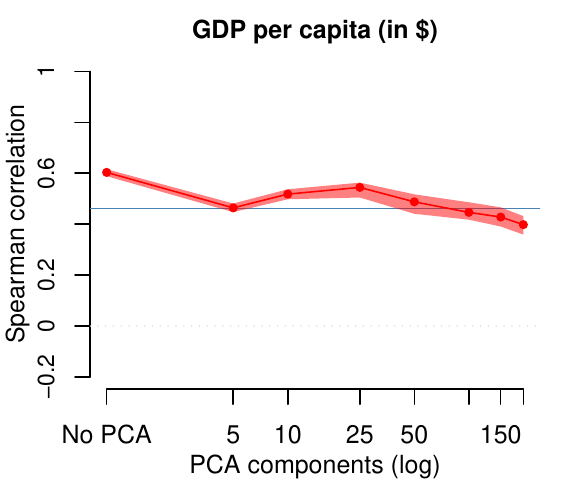}

\includegraphics[width=\ww\linewidth]{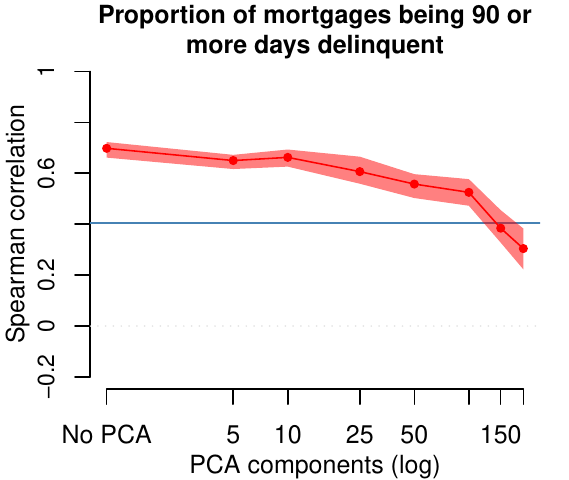}
\includegraphics[width=\ww\linewidth]{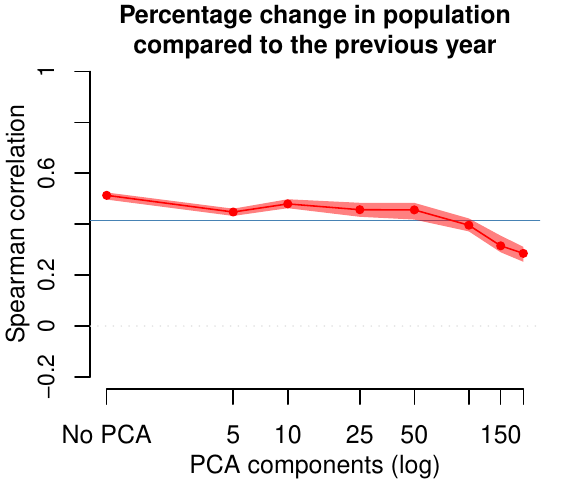}
\includegraphics[width=\ww\linewidth]{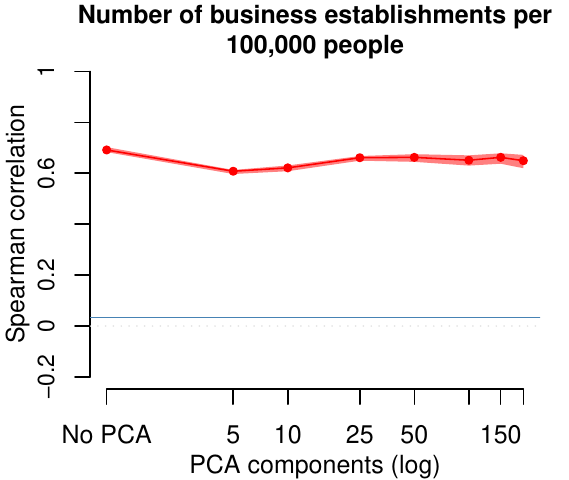}

\caption{Performance of the base model on the US counties dataset as a function of the number of PCA components. LME’s mean performance across 25 cross-validation iterations is shown as a line; the shaded band indicates the minimum–maximum range across iterations.}

\label{fig:performance_pca_US_spearman}
\end{figure}

\renewcommand\ww{.32}
\begin{figure}
\includegraphics[width=\ww\linewidth]{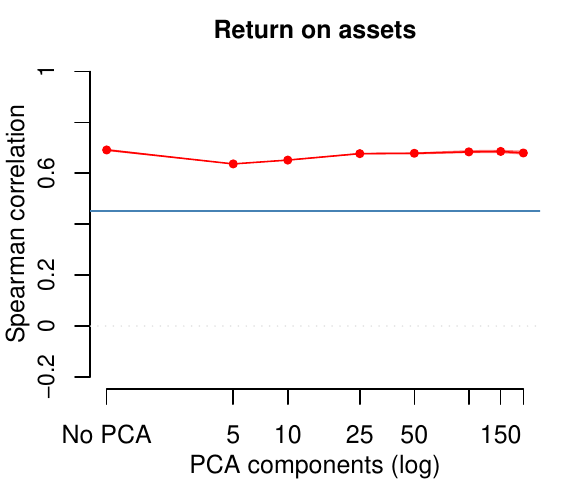}
\includegraphics[width=\ww\linewidth]{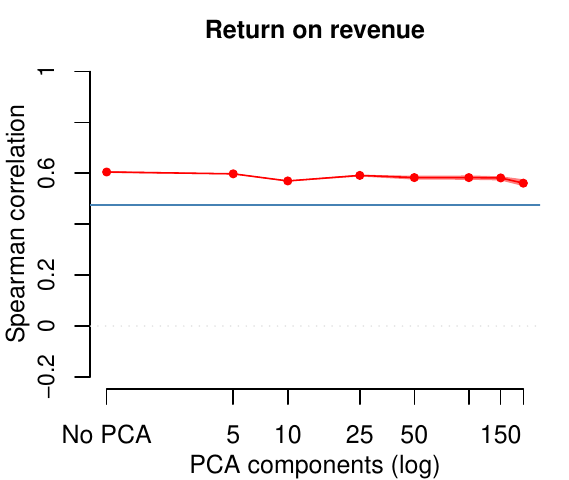}
\includegraphics[width=\ww\linewidth]{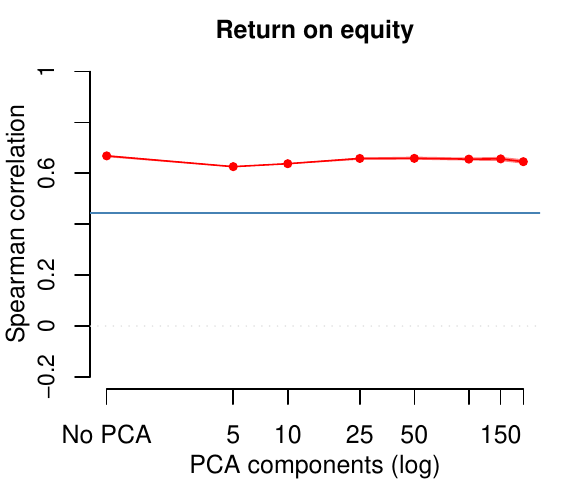}

\includegraphics[width=\ww\linewidth]{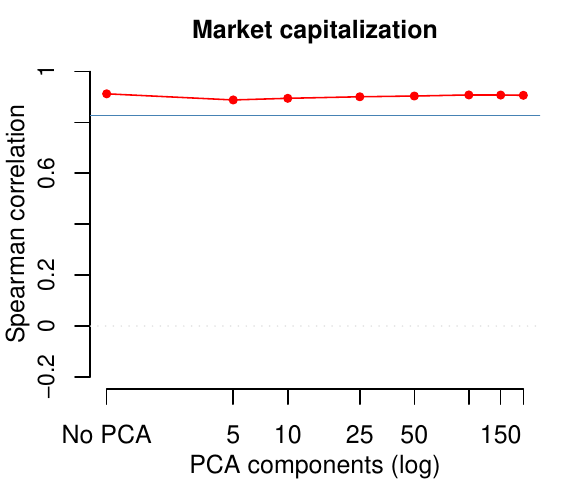}
\includegraphics[width=\ww\linewidth]{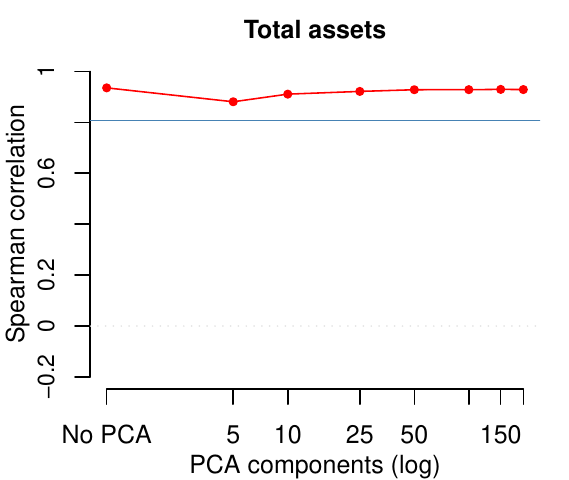}
\includegraphics[width=\ww\linewidth]{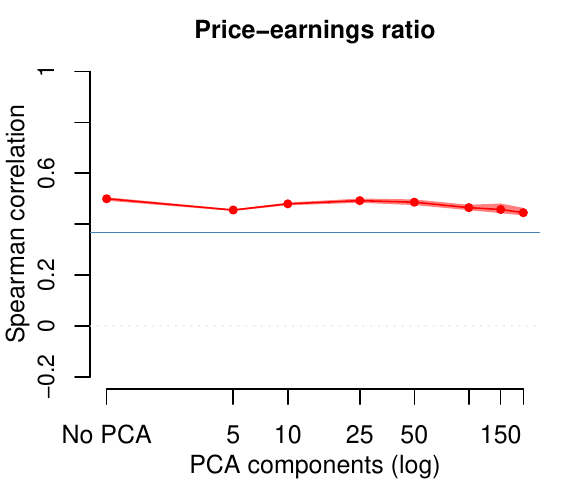}

\includegraphics[width=\ww\linewidth]{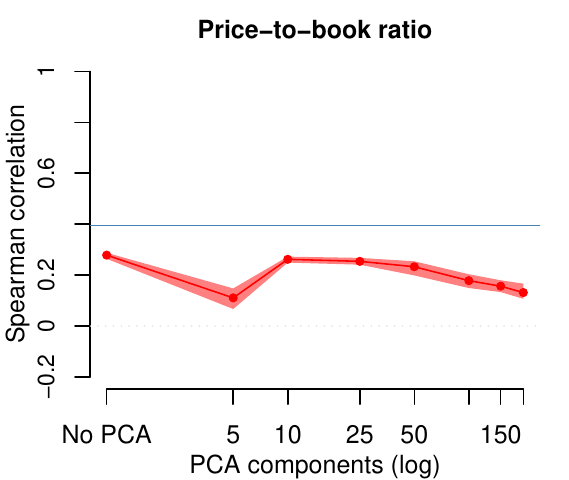}
\includegraphics[width=\ww\linewidth]{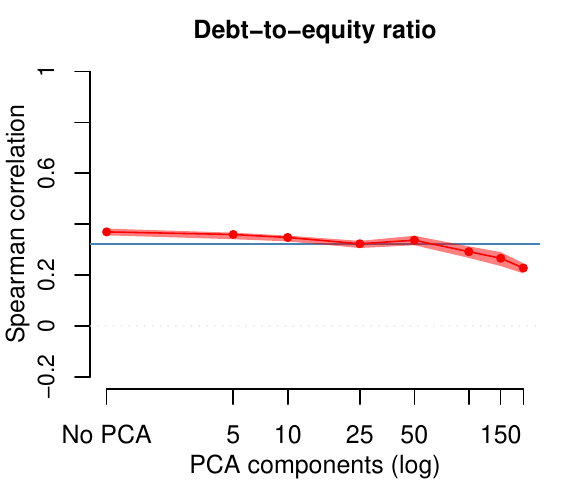}
\includegraphics[width=\ww\linewidth]{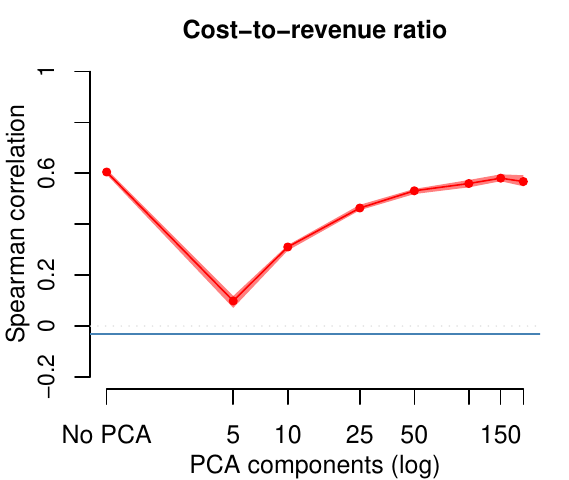}

\caption{Performance of the base model on the US-firms dataset as a function of the number of PCA components. LME’s mean performance across 25 cross-validation iterations is shown as a line; the shaded band indicates the minimum–maximum range across iterations.}
\label{fig:performance_pca_yahoo_spearman}
\end{figure}

\renewcommand\ww{.245}


\begin{figure}[!htb]
\begin{center}
\includegraphics[width=.8\linewidth]{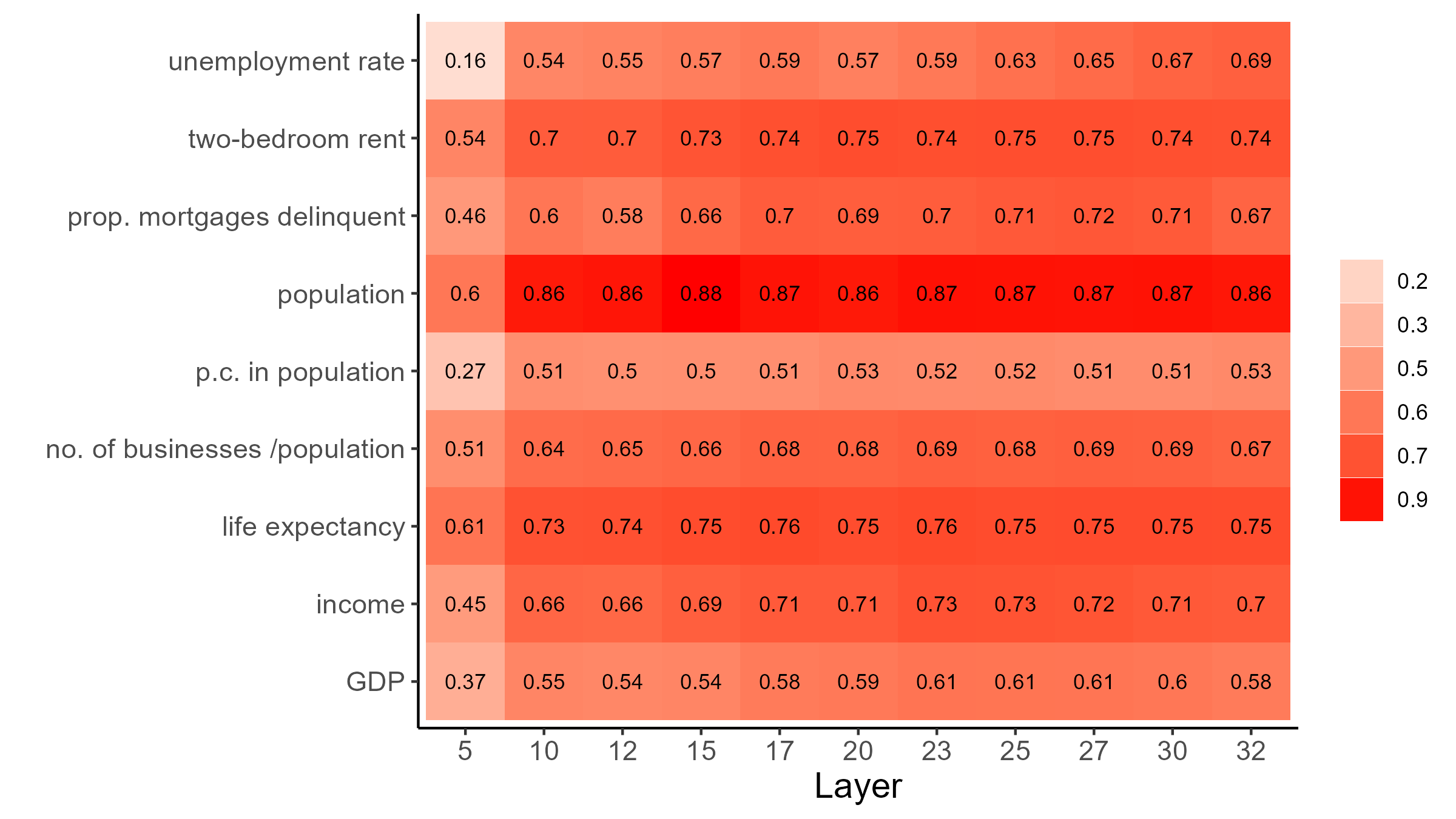}
\includegraphics[width=.8\linewidth]{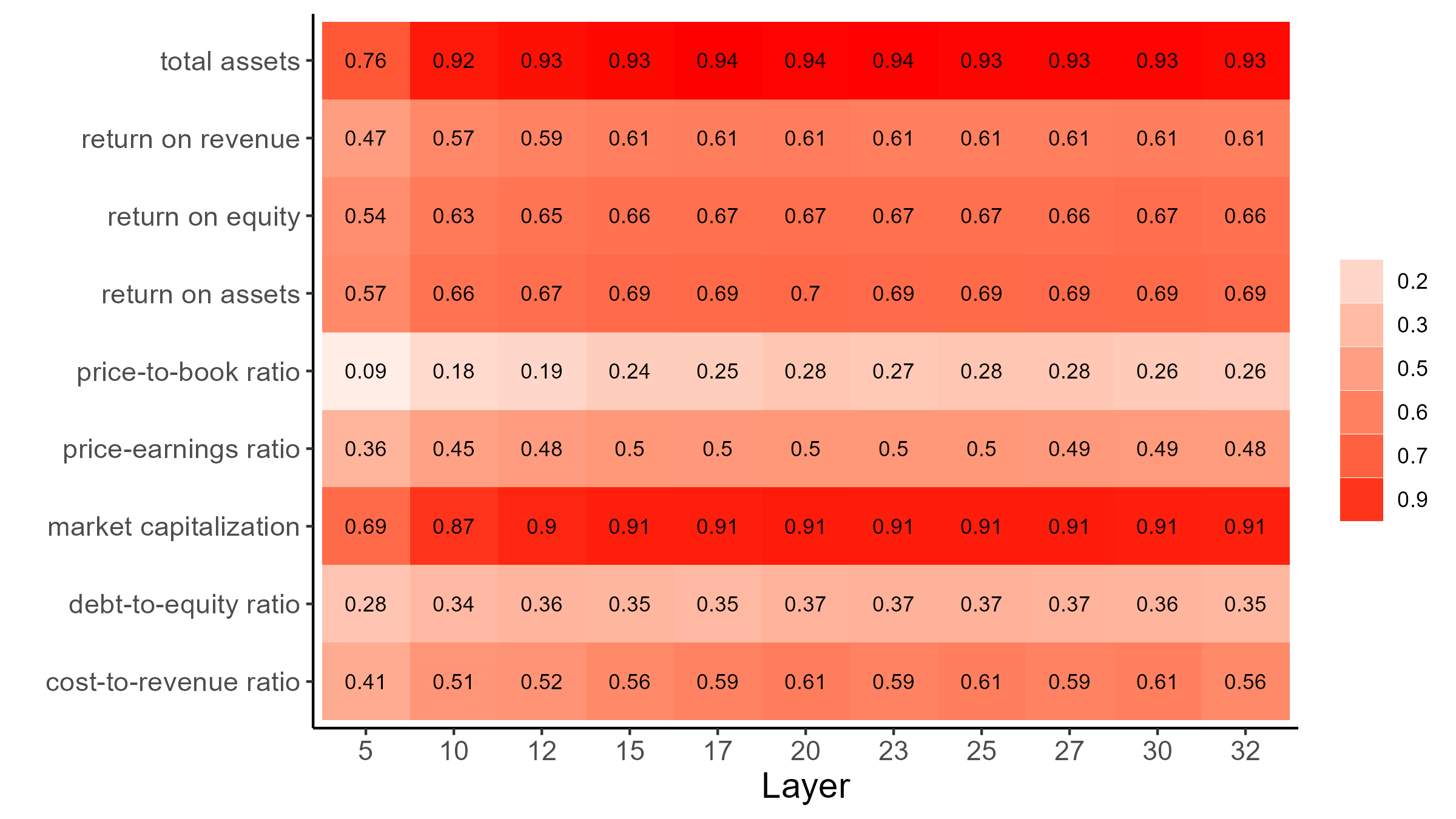}
\caption{Comparing LME performance when trained on different layers.}
\label{fig:by_layer}
\end{center}
\end{figure}

\renewcommand\ww{.32}

\begin{figure}
\includegraphics[width=\ww\linewidth]{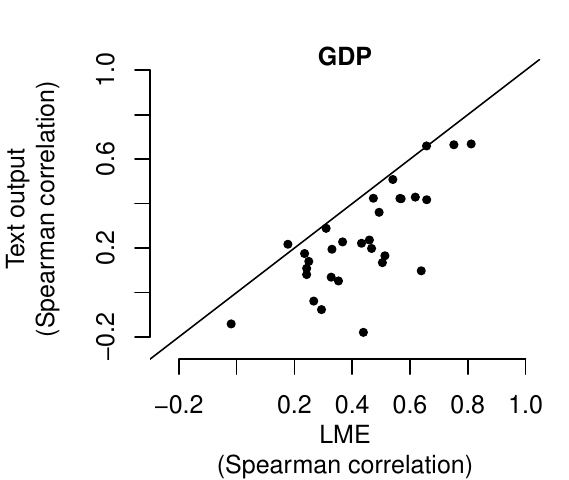} 
\includegraphics[width=\ww\linewidth]{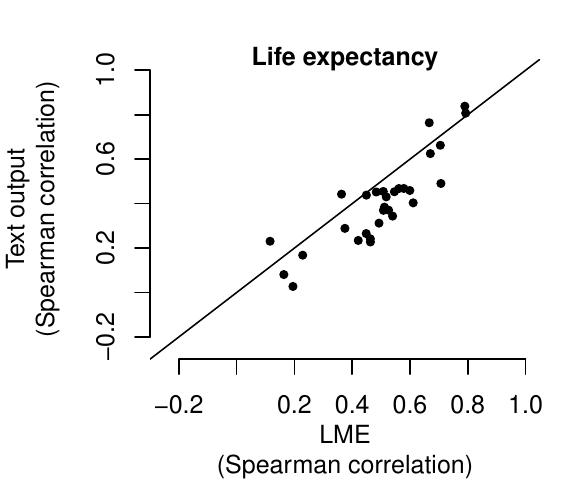} 
\includegraphics[width=\ww\linewidth]{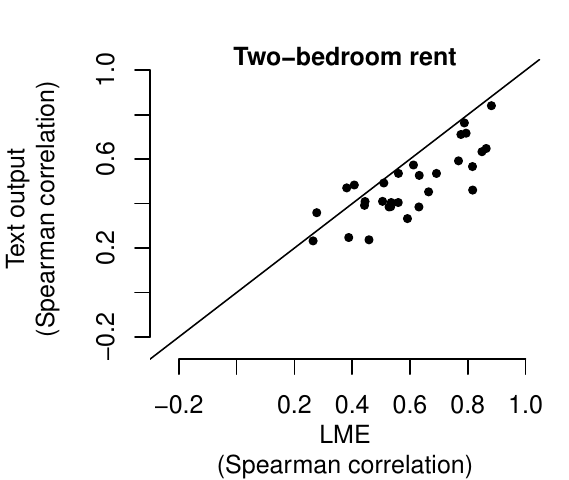} 
\includegraphics[width=\ww\linewidth]{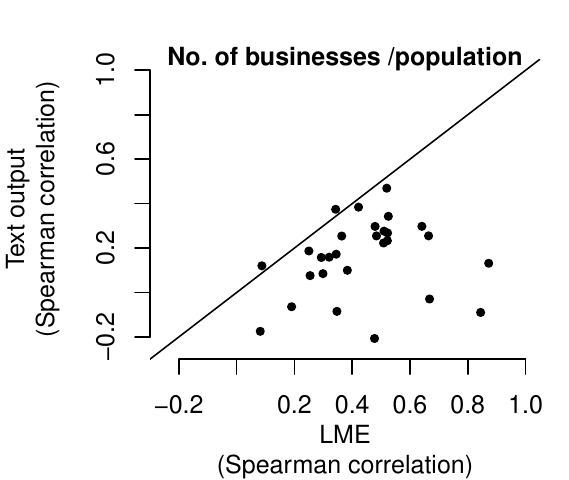} 
\includegraphics[width=\ww\linewidth]{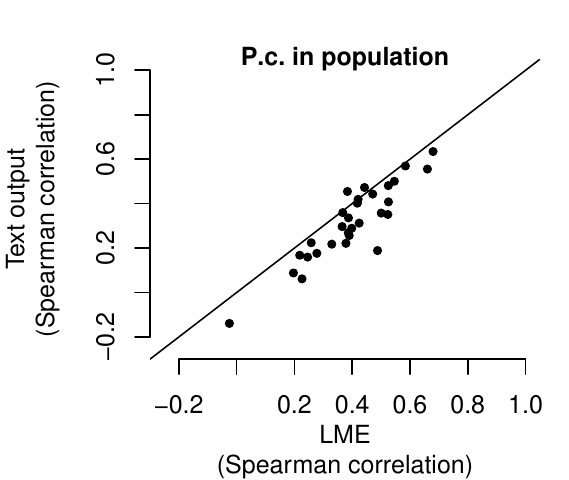} 
\includegraphics[width=\ww\linewidth]{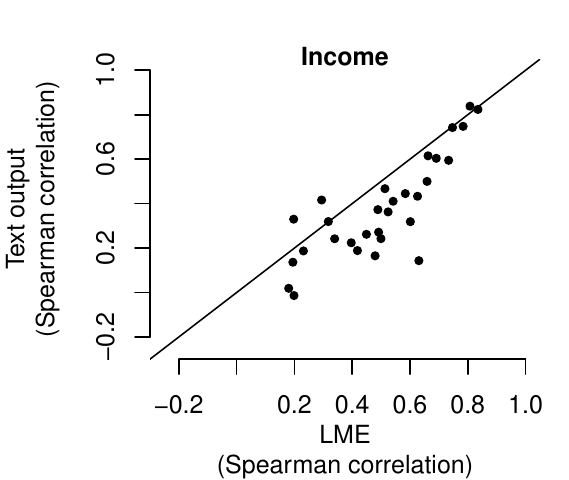} 
\includegraphics[width=\ww\linewidth]{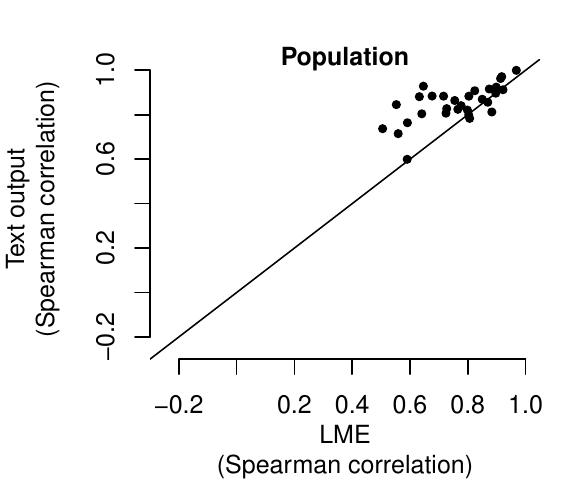} 
\includegraphics[width=\ww\linewidth]{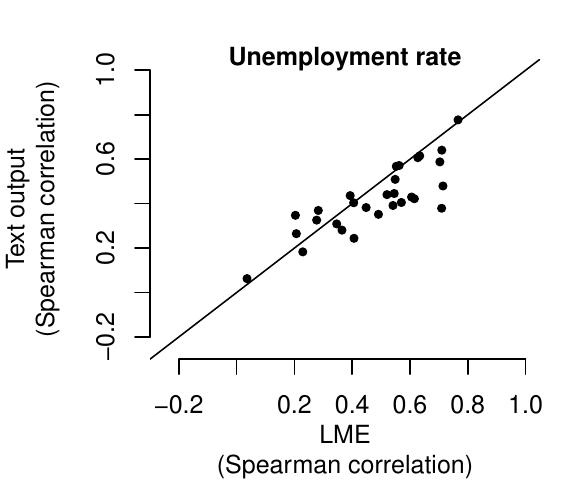} 
\caption{Comparison of performance between LME and text output within states with at least 50 counties.}
\label{fig:within_states}
\end{figure}

\begin{figure}
\includegraphics[width=1\linewidth]{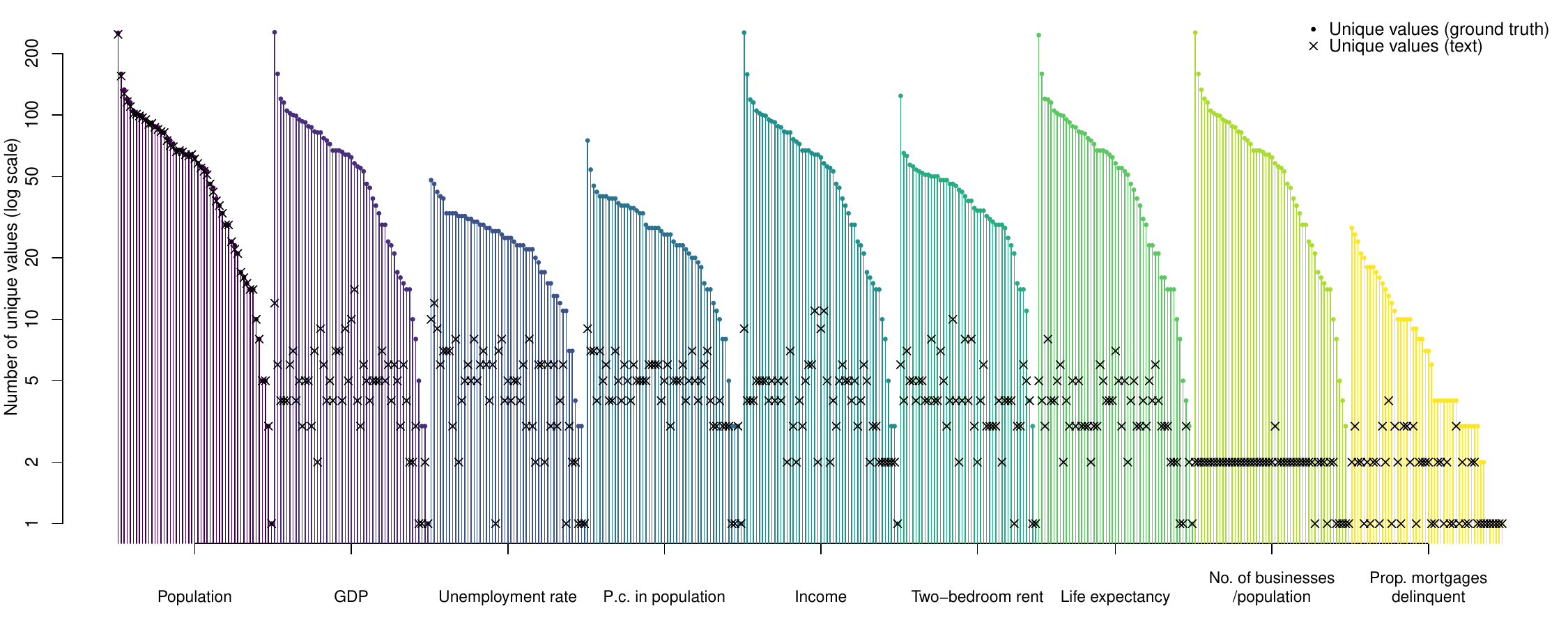}    
\caption{Number of unique ground truth and text output values for each state. Each vertical lines represents one state; states are ordered by the decreasing number of unique values observed.}
\label{fig:n_unique}
\end{figure}

%
%

\end{document}